\documentclass{article} 
\usepackage{iclr2026_conference,times}

\usepackage{amsmath,amsfonts,bm}

\def\eqref#1{equation~\ref{#1}}

\def\1{\bm{1}}

\DeclareMathAlphabet{\mathsfit}{\encodingdefault}{\sfdefault}{m}{sl}
\SetMathAlphabet{\mathsfit}{bold}{\encodingdefault}{\sfdefault}{bx}{n}

\usepackage{algorithm}
\usepackage{algorithmicx}
\usepackage{algpseudocode}
\usepackage[table]{xcolor}
\usepackage{xcolor}
\usepackage{url}
\usepackage{multirow}
\usepackage{array}
\usepackage{adjustbox}
\usepackage{arydshln}
\usepackage{wrapfig}
\usepackage{pifont}
\usepackage{caption}
\usepackage{booktabs}
\usepackage{float}
\usepackage{pifont}
\usepackage{makecell}

\definecolor{darkblue}{HTML}{1d1593}
\definecolor{highlight}{HTML}{C5B0CD}

\usepackage[colorlinks=true,linkcolor=red,citecolor=darkblue,urlcolor=darkblue]{hyperref}

\title{Proactive Reasoning-with-Retrieval \\ Framework for Medical Multimodal \\ Large Language Models}

\author{Lehan Wang, Yi Qin, Honglong Yang, Xiaomeng Li\thanks{Corresponding to Xiaomeng Li (eexmli@ust.hk).} \\ The Hong Kong University of Science and Technology}

\newcommand{\modelname}{\textsc{Med-RwR}}

\newcommand{\token}[1]{{\footnotesize\texttt{#1}}}
\newcommand{\tokensmall}[1]{{\fontsize{8.5pt}{9pt}\selectfont\texttt{#1}}}
\newcommand{\versus}{\textit{v.s.}}
\newcommand{\yes}{\ding{51}}
\newcommand{\no}{\ding{55}}

\iclrfinalcopy 
\begin{document}

\maketitle

\begin{abstract}
Incentivizing the reasoning ability of Multimodal Large Language Models (MLLMs) is essential for medical applications to transparently analyze medical scans and provide reliable diagnosis. However, existing medical MLLMs rely solely on internal knowledge during reasoning, leading to hallucinated reasoning and factual inaccuracies when encountering cases beyond their training scope.
Although recent Agentic Retrieval-Augmented Generation (RAG) methods elicit the medical model's proactive retrieval ability during reasoning, they are confined to unimodal LLMs, neglecting the crucial visual information during reasoning and retrieval. To this end, we propose the first \textbf{Multimodal Medical Reasoning-with-Retrieval framework, \modelname{}}, which proactively retrieves external knowledge by querying observed symptoms or domain-specific medical concepts during reasoning.
Specifically, we design a two-stage reinforcement learning strategy with tailored rewards that stimulate the model to leverage both visual diagnostic findings and textual clinical information for effective retrieval.
Building on this foundation, we further propose a \textbf{Confidence-Driven Image Re-retrieval (\textsc{CDIR})} method for test-time scaling when {low prediction confidence is detected.}
Evaluation on various public medical benchmarks demonstrates \modelname{}'s significant improvements over baseline models, proving the effectiveness of enhancing reasoning capabilities with external knowledge integration.
Furthermore, \modelname{} demonstrates remarkable generalizability to unfamiliar domains, evidenced by 8.8\% performance gain on our proposed EchoCardiography Benchmark (ECBench), despite the scarcity of echocardiography data in the training corpus. Our data, model, and codes will be made publicly available at \href{https://github.com/xmed-lab/Med-RwR}{https://github.com/xmed-lab/Med-RwR}.
\end{abstract}


\section{Introduction}
Reasoning has emerged as an essential capability in multimodal large language models (MLLMs), allowing the models to reveal the cognitive process of analyzing information, solving problems, and drawing conclusions. 
Existing works has employed techniques such as Long Chain-of-Thought Distillation~\citep{yao2024mulberry,dong2025insight,du2025virgo} and Reinforcement Learning~\citep{liu2025visual,meng2025mm,huang2025vision} to develop multimodal reasoning models, with recent efforts extending these approaches to healthcare~\citep{pan2025medvlm,su2025gmai,mu2025elicit}.

\begin{figure*}[t]
	\centering
	\includegraphics[width=1.0\textwidth]{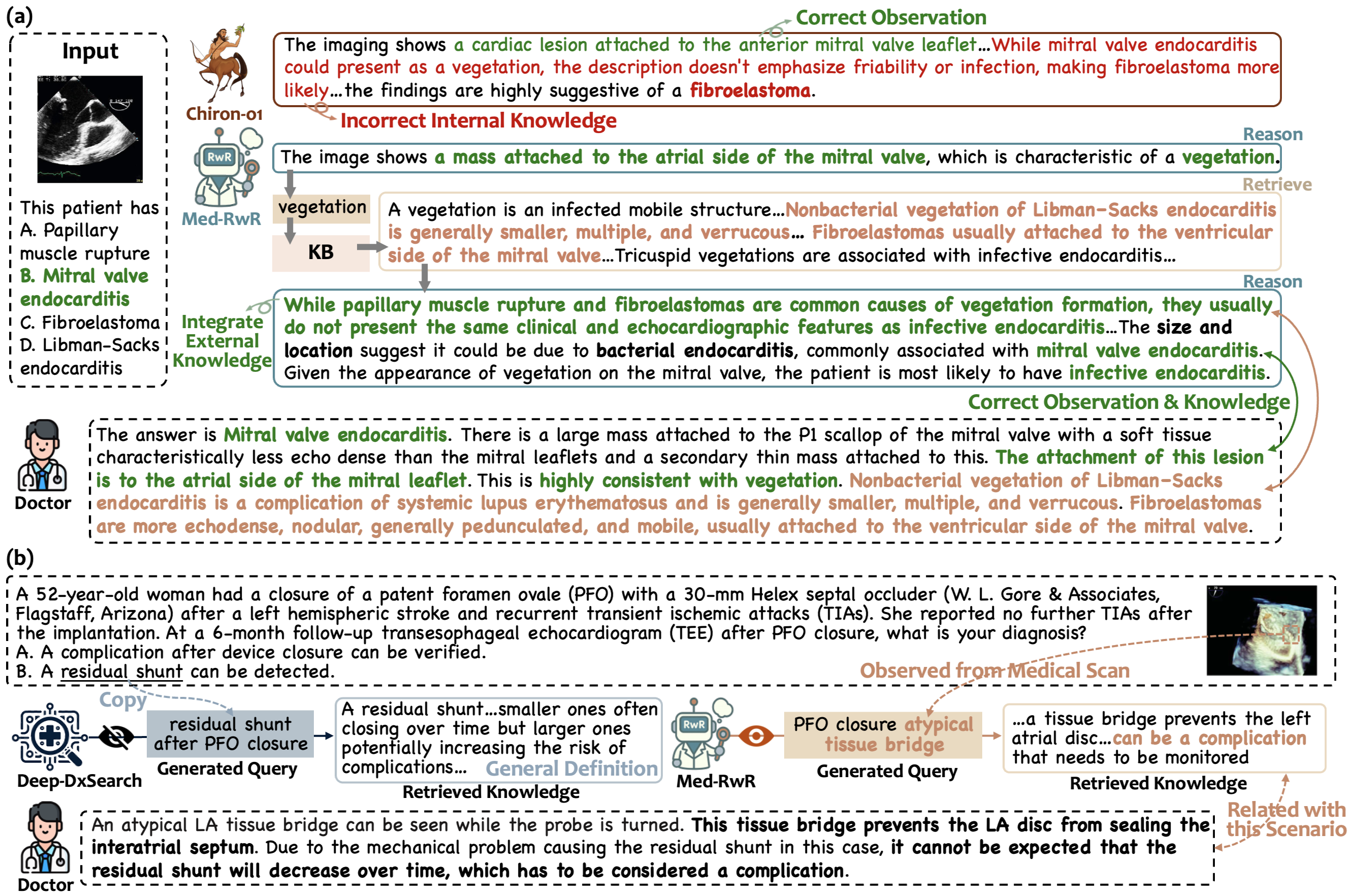}
	\caption{(a) Comparison between the generated outputs of a recent reasoning model, Chiron-o1~\citep{sun2025enhancing}, and the rollout process of our proposed Reasoning-with-Retrieval framework, which includes a trajectory of think, query and retrieve steps. Texts in \textcolor[HTML]{3B7D23}{green} denote the correct statements, \textcolor[HTML]{C00000}{red} indicates erroneous outputs, and \textcolor[HTML]{C68767}{brown} annotates the authoritative knowledge. (b) Comparison between the effectiveness of retrieval between text-only agentic RAG system for diagnosis, Deep-DxSearch~\citep{zheng2025end}, and our proposed multimodal \modelname{}. 
    \vspace{-1em}
    }
	\label{fig:intro}
\end{figure*}

However, ensuring the factual reliability of the reasoning process remains a significant challenge. 
Unreliable reasoning processes are inclined to propagate erroneous intermediate conclusions, resulting in incorrect decisions despite following solid logical structures.
As shown in Figure~\ref{fig:intro} (a), a recent medical reasoning model, Chiron-o1~\citep{sun2025enhancing}, demonstrates logical reasoning flow but excludes the correct answer by referring to invalid diagnostic criteria (friability, infection) that are not distinguishing features. 
This occurs because existing medical reasoning models solely rely on \textbf{\textit{internal knowledge}} during inference, making them prone to generate non-factual contents when encountering cases beyond their training scope. {This in turn leads to} misdiagnoses and reduces clinical trustworthiness, as exemplified in Figure~\ref{fig:intro} (a), where the model mistakenly identifies the condition as ``fibroelastoma" instead of the ground-truth answer due to reliance on false evidence.
Consequently, medical AI models should ground their reasoning in clinically authoritative evidence, rather than static and memorized knowledge alone. 

While Retrieval-Augmented Generation (RAG) could expose the models to clinical evidence from external medical databases, current medical RAG models~\citep{jeong2024improving,ong2024surgeryllm,zhao2025medrag} are constrained by static retrieval. 
To mitigate this issue, Agentic Search~\citep{jin2025search,wu2025mmsearch} has emerged as a promising approach, and \cite{ding2025promed,zheng2025end,yu2025medreseacher} have adapted this framework to the medical domain, encouraging models to actively integrate \textbf{\textit{external evidence}} from the knowledge base during reasoning. Nevertheless, they focus primarily on Large Language Models fed with only textual input. 
This is insufficient in medical diagnosis since the rich visual context may be overlooked during the proactive query generation and knowledge retrieval processes, thus fetching general definitions and failing to correspond to specific evidence in the scans. For instance, the text-only agentic RAG system, Deep-DxSearch~\citep{zheng2025end}, can only execute the retrieve action based on textual hints without images as inputs. Figure~\ref{fig:intro} (b) demonstrates that Deep-DxSearch directly copies the query from the original question, resulting in retrieval of a general medical definition that lacks the discriminative features for differential diagnosis. This challenge arises because retrieving relevant knowledge requires leveraging key observations from the image, which remains largely underexplored.

{To address this limitation, we propose \textbf{\modelname{}}, a multimodal \textbf{Med}ical \textbf{R}easoning-\textbf{w}ith-\textbf{R}etrieval pipeline, which incentivizes the medical MLLM to actively retrieve factual evidence during reasoning. 
Our core idea is to mirror how clinicians reason to diagnose: observing the medical scans while cross-referencing textual protocols and guidelines, then making final conclusions based on evidence. 
Specifically, we first implement a Cluster-Reconstruct-Stratify data generation process to curate a controllable environment, comprising a multimodal training dataset and a knowledge base.
Subsequently, we develop a two-stage reinforcement learning (RL) framework guided by a tailored reward design aware of both visual findings and textual information, thereby ensuring the retrieved knowledge is aligned with both modalities. To augment inadequate knowledge retrieval, the trained framework further supports \textbf{C}onfidence-\textbf{D}riven \textbf{I}mage \textbf{R}e-retrieval (\textbf{\textsc{CDIR}}) for test-time scaling, where it can {refer to} similar multimodal cases when making less confident decisions. As illustrated in Figure~\ref{fig:intro}, our model first excludes the differential diseases due to incompatible size and location by querying the underlying cause of the observed phenomenon, and then confirms the correct answer via retrieved diagnosis criteria.

}

To demonstrate the effectiveness of our proposed method, we conduct extensive experiments on public benchmarks, and achieve an improvement of 5.1\% on MedXpertQA-MM, 9.7\% on MMMU-H\&M, and 12.3\% on MMMU-Pro-H\&M, respectively, compared with the base model.
Furthermore, {one of the representative advantages of \modelname{} is its \textit{\textbf{generalizability to unfamiliar domains}} achieved by leveraging the strong reasoning-with-retrieval capacity.
To demonstrate this, we evaluate the model on echocardiography, a highly specialized medical domain that poses challenges for adaptation due to both data scarcity (less than 2\% of the training corpus as shown in Figure~\ref{fig:data-distribution}) and high clinical expertise requirements. We construct EchoCardiography Benchmark (ECBench) from authoritative practice collections of multiple difficulty levels, benchmarking the model on the underrepresented multimodal echocardiography data.} 
Our method achieves performance gains of at least 8\% over other state-of-the-art methods on ECBench, demonstrating its flexibility to incorporate domain-specific external knowledge for adaptation.

To summarize, we establish \textbf{the first comprehensive multimodal Medical Reasoning-with-Retrieval framework, \modelname{}}, encompassing contributions as follows:



\textbullet \ \ We develop an environment with a curated multimodal dataset and a specialized knowledge base to train the medical MLLM to \textit{actively ground the reasoning process in external reliable sources}, rather than depending exclusively on internal knowledge.

\textbullet \ \ We develop {a two-stage reinforcement learning (RL) strategy with tailored rewards} specifically designed to stimulate the medical MLLM to \textit{be aware of both visual findings and textual contexts in retrieval} for more relevant and accurate information.

\textbullet \ \ We demonstrate the {effectiveness} of our approach in \textit{enhancing medical reasoning and understanding} through extensive evaluation, achieving significant improvements on public medical multimodal benchmarks. To evaluate \textit{generalizability to scarce medical domains}, we collect ECBench, a multimodal echocardiography benchmark. Our method demonstrates remarkable adaptation to this specialty through leveraging external expertise despite limited domain-specific training data.

\vspace{-1em}
\section{Related Works}
\subsection{Medical Multimodal Large Language Models}
Current advances in MLLMs have accelerated their application in medical settings to assist diagnostic procedures~\citep{wu2023towards,bannur2024maira,zhang2024generalist,wang2024interpretable,yang2025multi}. 
However, these models directly provide diagnostic outputs without detailed explanations, making it difficult for clinicians to interpret and follow. This lack of transparency has driven the development of medical reasoning models, which explicitly output the thinking process for decision-making. Several works focus on curating reasoning chains to teach medical models to think step-by-step~\citep{chen2024huatuogpt,huang2025o1,wu2025medreason}, 
and recent efforts have applied reinforcement learning to iteratively improve reasoning quality through reward-based training~\citep{xu2025medground,su2025gmai,liu2025x,mu2025elicit}. Despite these improvements, both approaches rely heavily on the models' internal knowledge in the inference stage, which can generate reasoning processes that contain unreliable medical information especially when encountering data beyond training scope, potentially compromising clinical decision-making.
\vspace{-0.5em}
\subsection{Agentic Search and Retrieval}
Researchers have explored integrating retrieval into the reasoning process to enable mutual enhancement, where the retrieved information can support reasoning while reasoning helps generate more effective queries.
Recent works have introduced Reinforcement Learning (RL) to stimulate {this process.} 
Search-R1~\citep{jin2025search} and R1-Searcher~\citep{song2025r1} proposed search-augmented RL training strategies to enhance retrieval-driven reasoning and decision-making. 
In the multimodal domain, Visual-ARFT~\citep{liu2025visual2} equips MLLMs with agentic capabilities of text-based searching. MMSearch-R1~\citep{wu2025mmsearch} allows the model to perform adaptive multimodal retrieval from real-world Internet sources. 
Despite these advances, there remains a critical gap in exploring multimodal reasoning with adaptive retrieval based on both imaging and text in the medical domain, where the retrieved knowledge often supports differential diagnosis reasoning among multiple hypotheses, rather than seeking a single correct answer. This requires the model to be aware of anatomical structures and pathological patterns {in the medical scans}, a capability that general text-based agentic search models lack. 

\vspace{-1em}
\section{Methodology}
In this section, we elaborate on the proactive multimodal reasoning-with-retrieval framework. 
We first curate a controllable environment with a complex multimodal training dataset and a specialized knowledge base (\S\ref{sec:data}).
With the training environment prepared, we design a two-stage reinforcement learning (RL) strategy with tailored rewards to incentivize external knowledge retrieval during reasoning (\S\ref{sec:train}). To augment inadequate information retrieval during inference, we propose a confidence-driven image re-retrieval scheme for test-time computational scaling (\S\ref{sec:infer}).
\subsection{Reasoning-with-Retrieval Environment}
\label{sec:data}
Our training environment for facilitating reasoning-with-retrieval ability comprises a multimodal training dataset and a knowledge base carefully aligned with its scope.
To construct a multimodal dataset with distinct difficulty levels for reinforcement learning, we employ a Cluster-Reconstruct-Stratify pipeline using PubMedVision~\citep{chen2024huatuogpt-2} as the data source. 
\textbf{Cluster}: Firstly, we perform K-means clustering to group all image features into 2,000 clusters to balance between diversity and efficiency
, and sample representative instances near cluster centers to ensure coverage of diverse visual patterns. 
\textbf{Reconstruct}: We input selected images with original captions into GPT-4o to extract background knowledge, observation, analysis and conclusion. Then, we further prompt it to reformulate multi-choice questions asking for conclusions based on the background without revealing any visual observation or analysis. Illustrative examples are presented in Figure~\ref{fig:train-sample}.
\textbf{Stratify}: To assess question difficulty, we employ two MLLMs, QwenVL2.5-7B~\citep{bai2025qwen2} and InternVL3-8B~\citep{zhu2025internvl3} to generate 10 responses for each question and measure accuracy. We remove trivial questions with correct answers in all iterations, and stratify the remaining questions into easy and difficult categories based on accuracy thresholds.
Finally, we obtain approximately 6,500 multimodal question-answer pairs of different difficulty levels for progressive curriculum training. Dataset statistics are in Appendix~\ref{sec:data-stat}. 
To support the retrieval action of our framework, we construct a medical knowledge base by extracting medical knowledge from the analytical contents generated during the training data construction process, as detailed in Appendix~\ref{sec:kb-detail}. 
\subsection{Multimodal Medical Reasoning-with-Retrieval Framework}
\label{sec:train}

\begin{figure*}[t]
	\centering
	\includegraphics[width=1.0\textwidth]{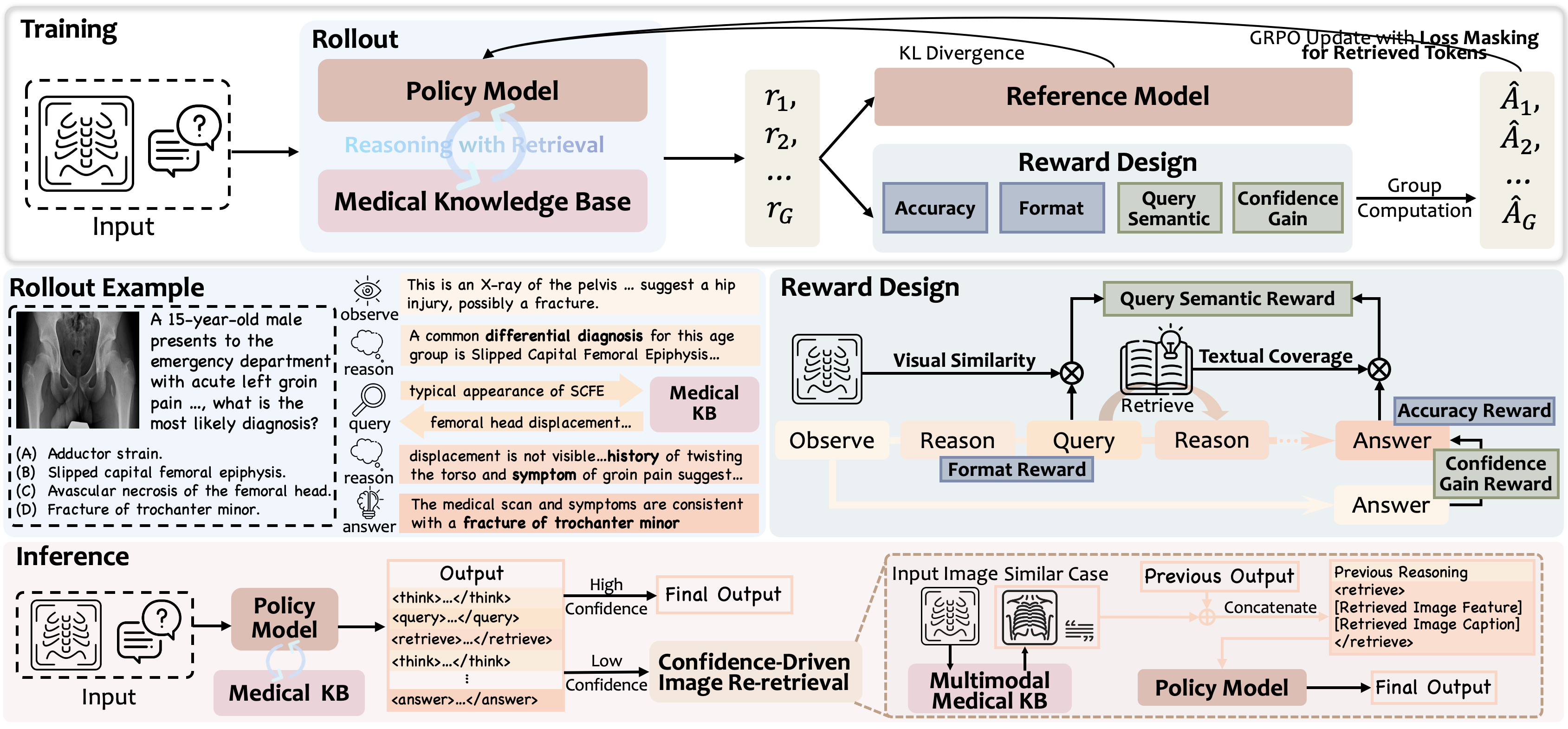}
	\caption{Overall Illustration of the Reasoning-with-Retrieval framework. During training, the model learns to generate reasoning steps while actively retrieving from a medical knowledge base. During inference, the Confidence-Driven Image Re-retrieval scheme would be triggered to augment less informative retrieval when low confidence is detected.}
	\label{fig:frame}
\end{figure*}

\subsubsection{Problem Formulation}
\label{sec:problem}


We formulate medical visual question answering as a reasoning-with-retrieval framework. 
Specifically, the MLLM is modeled as the policy model $\pi_\theta$($\cdot\mid x$), which {performs autoregressive prediction} given input $x=\{i,t\}$ consisting of medical images $i$ and text prompts $t$. 
The model $\pi_\theta$ first outputs reasoning steps $h$ based on the multimodal input $x$ before generating the final response $y$. During reasoning, the model can decide to generate a query $q$ to retrieve relevant information $k \in \mathcal{K}$ from an external medical knowledge database $\mathcal{K}$. The retrieved knowledge $k$ and the reasoning history $h$ are then incorporated into the model's context, enabling the model to continue generation with enriched information, $\pi_\theta(\cdot\mid x,h,k)$, until reaching a final answer.


\subsubsection{Reward Design}
\label{sec:reward}
To equip our proposed framework with proactive reasoning-with-retrieval ability, we design four rewards to ensure that the model learns effective retrieval strategies to seek highly relevant knowledge that aligns with the multimodal inputs. 


\textbf{Retrieval Format Reward.}
We first define the format reward to ensure structured output generation. The model is expected to enclose the thinking process within \token{<think>} and \token{</think>} and retrieval query within \token{<query>} and \token{</query>}. The retrieval system will search in the database $\mathcal{K}$ based on the query and return the matched information within \token{<retrieve>} and \token{</retrieve>}. 
The final answer is enclosed between \token{<answer>} and \token{</answer>}. The format reward $\sigma_f$ equals 1 if the output includes a thinking process, with an additional 1 when retrieval is activated and leads to a correct answer, and 0 otherwise.


\textbf{Accuracy Reward.}
The accuracy reward directly evaluates whether the model's output is correct. The answer choice is extracted from the generation and compared against the ground truth. The accuracy reward $\sigma_a$ is set to 1 for correct choices and 0 for incorrect ones.

\textbf{Query Semantic Reward.} 
{We propose the Query Semantic Reward to drive the model to generate queries with enhanced semantic alignment between both the text prompt and visual content. This could improve the contextual relevance of the retrieval process, contributing to enhanced response accuracy. The reward consists of two components to separately address textual and visual alignment.

The first part of the reward promotes the medical-specific textual semantic alignment among the generated query, retrieved contents, and ground-truth content. 
Specifically, we identify medical entities from them using a named-entity recognition (NER) model~\citep{Kraljevic2021-ln} pretrained to detect clinical concepts within the Unified Medical Language System (UMLS). The reward is then calculated as the proportion of overlap entities, thereby incentivizing the model to generate queries that drive the reasoning process toward the ground truth. We give the formal formulation of this reward as Eqn.~\ref{eq:reward1} below, where $S_Q, S_K,S_G$ denote the medical entity set of generated query, retrieved information, and ground truth content.
\begin{equation}
    \sigma_{q_\text{text}}=\frac{|S_Q \cap S_G|}{|S_Q|}+\frac{|S_K \cap S_G|}{|S_K|} ,
    \label{eq:reward1}
\end{equation}
The second part of the reward enhances the visual semantic alignment between the generated query and the input image. This design prevents the model from learning shortcuts~\citep{xia2025visionary}, i.e., querying solely based on the textual prompt while ignoring the visual information, and improves the visual relevance of the retrieved knowledge. Specifically, it maximizes the semantic similarity between the generated query and the input image within the shared embedding space, with formal formulation given as Eqn.~\ref{eq:reward2}, where $i,t$ denote the visual and textual input, and $f_{image},f_{text}$ are the pretrained image and text encoders from BiomedCLIP~\citep{zhang2023biomedclip}. 
\begin{equation}
    \sigma_{q_\text{image}}=\frac{f_\text{image}(i) \cdot f_\text{text}(t)}{||f_\text{image}(i)|| \ ||f_\text{text}(t)||} , \label{eq:reward2}
\end{equation}
Through additional visual alignment, the model is encouraged to leverage visual features during query formulation. The final query semantic reward is a linear combination of both Eqn.~\ref{eq:reward1} and Eqn.~\ref{eq:reward2}, denoted as $\sigma_q=\sigma_{q_\text{text}}+\sigma_{q_\text{image}}$.
}

\textbf{Confidence Gain Reward.} 
\label{sec:cgr}
{To ensure that the retrieval process substantively contributes to reaching the accurate response, we establish a connection between the retrieval mechanism and the model's confidence in providing the correct answer. Intuitively, a model supported by the retrieved knowledge should exhibit higher confidence in predicting the correct answer compared to when retrieval is absent. Building on this, we propose the confidence gain reward, which quantifies the improvement in the model's confidence between its output with retrieved content $y$ and its output without retrieved content $y'$. Specifically, let the token position of \token{<answer>} be denoted as $\tau$. At the position $\tau+1$, the model generates the answer to a multiple-choice question. To compute the confidence gain, we measure the difference in the predicted log probabilities of the ground-truth answer token before and after the retrieval process.
Eqn.~\ref{eq:reward3} gives the formulation of confidence gain reward, where $h$ denotes the previous reasoning history, $k$ denotes the retrieved information.}
\begin{equation}
    \sigma_{c} = \log \frac{\pi_{\theta}((y')_{\tau'+1}^{\text{answer}}\mid x,y'_{\leq\tau'},h,k)}{\pi_{\theta}(y_{\tau+1}^{\text{answer}}\mid x,y_{\leq\tau})}  ,  \label{eq:reward3}
\end{equation}
\textbf{Overall Rewards.} Finally, we rescale each reward to ensure balanced contribution using weights $w_f=1, w_a=5, w_q=0.4, w_c=5,$ demonstrated in Eqn.~\ref{eq:all-reward}. 
\begin{equation}
    \sigma=w_f\sigma_f+w_a\sigma_a+w_q\sigma_q+w_c\sigma_c ,  \label{eq:all-reward}
\end{equation}
\vspace{-2em}
\subsubsection{Model Training}

\textbf{Group Relative Policy Optimization (GRPO).} Following \cite{guo2025deepseek}, we employ GRPO algorithm with the proposed rewards to steer the model's exploration in reasoning and retrieval. Specifically, for each training instance, we sample a group of rollouts $\{r_i\}_{i=1}^G$ from the policy model $\pi_{\theta_\text{old}}$ and acquire the reward $\sigma_i$. The advantage $\hat{A}_{i,t}$ 
is estimated as the standard score of the group rewards. 
GRPO objective is calculated as shown in Eqn.~\ref{eq:grpo_objective}:
\begin{align}
\mathcal{J}_{\text{GRPO}}(\theta) = & \mathbb{E}_{x, \{r_{i,s}\}_{i=1,s=1}^{G,S} \sim \pi_{\theta_{\text{old}}}(\cdot|x,h,k)} \Bigg\{ \frac{1}{GS|r_{i,s}|} \sum_{i=1}^{G} \sum_{s=1}^{S} \sum_{t=1}^{|r_{i,s}|} \min \left[ \rho_{i,s,t} \hat{A}_{i,s,t}, \text{clip}  (\rho_{i,s,t}, 1-\epsilon, \right. \nonumber \\ & \hspace{-1cm} \left. 1+\epsilon) \hat{A}_{i,s,t} \right] 
- \beta \mathbb{D}_{\text{KL}}[\pi_\theta(\cdot|x,r_{i,<s,<t},h_{i,<s},k_{i,<s}) \| \pi_{\theta_{\text{ref}}}(\cdot|x,r_{i,<s,<t},h_{i,<s},k_{i,<s})] \Bigg\}, \label{eq:grpo_objective}
\end{align}
where $\rho_{i,t}=\frac{\pi_\theta(r_{i,s,t} | x, r_{i,<s,<t},h_i,k_i)}{\pi_{\theta_{\text{old}}}(r_{i,t} | x, r_{i,<t},h_i,k_i)}$, $h_i,k_i$ are the reasoning history and retrieved knowledge for rollout $i$. We adopt the Token-level Policy Gradient Loss~\citep{yu2025dapo} to mitigate length bias, and tokens from the retrieved knowledge are masked during loss calculation.

\textbf{Two-Stage Reinforcement Learning Training.} Motivated by the empirical evidence that multimodal medical reasoning requires first establishing foundational reasoning skills before integrating additional visual information~\citep{peng2025lmmr1empowering3blmms}, we utilize a two-stage Reinforcement Learning training strategy.
In the first stage, we train the model with text-only question-answer pairs sampled from MedQA-USMLE~\citep{jin2021disease} and MedMCQA~\citep{pal2022medmcqa}. 
We apply accuracy and format rewards to instill the model's fundamental medical reasoning-with-retrieval capabilities.
In the second stage, we extend to multimodal training with our constructed data.
All four reward types are incorporated to further encourage effective knowledge retrieval during reasoning.
This two-stage training scheme allows the model to first leverage its medical reasoning-with-retrieval competencies and gradually adapt to medical imaging. 

\subsection{Confidence-Driven Test-Time Computation Scaling via Image Re-retrieval}
\label{sec:infer}
With our proposed confidence-based reward modeling (Eqn.~\ref{eq:reward3}), the model is expected to output high confidence response through effective knowledge retrieval. 
When the model's final decision confidence remains low despite retrieval actions, it implies insufficient information in the knowledge base for the given case. 
This mirrors real clinical practice where clinicians find guidelines inadequate for confident decision-making. Under this circumstance, they also consult multimodal records from similar patient cases.
{Motivated by this, our model unlocks the capability of \textbf{C}onfidence-\textbf{D}riven test-time computation scaling via \textbf{I}mage \textbf{R}e-retrieval (\textsc{\textbf{CDIR}}) to enhance the diagnostic accuracy. Specifically, Eqn.~\ref{eq:reward3} employs answer probability to estimate output confidence, which correlates with response correctness. We first extract the confidence $\eta$ of the generated answer token $\hat{y}_{\tau+1}^\text{answer}$ using Eqn.~\ref{eq:cdir}:
\begin{equation}
    \eta=\pi_{\theta}(\hat{y}_{\tau+1}^{\text{answer}}\mid x,\hat{y}_{\leq\tau},h,k), \label{eq:cdir}
\end{equation}
where $\hat{y}$ is the output before image re-retrieval and $\hat{y}_{\tau+1}^{\text{answer}}$ denotes the probability of the predicted answer option at token position $\tau+1$ similar to the position at \S\ref{sec:cgr}. 
When the confidence score $\eta$ is below the threshold $\lambda$, i.e., $\eta < \lambda$, the model is more likely to yield an incorrect conclusion, indicating the need for multi-faceted knowledge to rectify the response.
Hence, we enable the model to re-retrieve analogous cases by computing image feature similarity between the input image and candidate images from the multimodal database $\mathcal{D}$ (See Appendix~\ref{sec:multi-kb} for details), where each image is paired with a corresponding clinical description. 
The retrieved image-caption pairs are then integrated into existing contexts to re-generate the response as in Eqn.~\ref{eq:final}.
\begin{equation}
    y\sim \pi_\theta(\cdot\mid i,t,h,k'), \text{ where } k' = 
    \begin{cases}
        k, & \eta \geq \lambda \\
        k \cup k_{\text{sim}}(i, \mathcal{D}), &\eta < \lambda
    \end{cases}
    \label{eq:final}
\end{equation}
where $k_{\text{sim}}(i, \mathcal{D})$ denotes the retrieved set of similar image-caption pairs from multimodal corpus $\mathcal{D}$ based on similarity with the input image $i$. Empirically, we set $\lambda=0.8$ in the experiment.}

\vspace{-1em}
\section{Experiments}
\vspace{-0.5em}
\subsection{Experimental Settings}
\newcommand{\reproduce}[1]{\textcolor[gray]{0.5}{#1}}

\textbf{Public Benchmarks.} We evaluate our model on three multimodal medical benchmarks, MedXpertQA-MM~\citep{zuo2025medxpertqa}, MMMU-H\&M~\citep{yue2024mmmu}, and MMMU-Pro-H\&M~\citep{yue2024mmmu2}. MedXpertQA-MM is a challenging benchmark collected from medical exams and
textbooks. MMMU-H\&M and MMMU-Pro-H\&M are subsets of Health and Medicine domain derived from MMMU and MMMU-Pro benchmarks. These benchmarks target the medical MLLMs' knowledge understanding and complex reasoning abilities. 

\textbf{Specialized Domain Benchmark.} To further demonstrate our model's reasoning-with-retrieval ability to generalize to unfamiliar domains, we propose a multimodal EchoCardiography Benchmark, ECBench, collected from clinical practice books. The benchmark contains 824 questions of echocardiogram interpretation in clinical scenarios.
See Appendix~\ref{sec:ecbench} for curation details. 


\textbf{Baseline Methods.} We conduct extensive comparisons against three categories of models: (1) Agentic Search MLLMs equipped with agentic abilities to dynamically search external knowledge sources for enhanced reasoning capabilities. (2) Generalist Medical MLLM capable of perceiving diverse medical modalities and resolving versatile tasks; (3) Reasoning Medical MLLMs optimized for complex reasoning.
Further implementation details can be found in Appendix~\ref{sec:impl}.

\vspace{-0.5em}
\subsection{Main Results}

\begin{table}[t]
    \caption{Performance on public multimodal medical benchmarks assessing medical MLLMs' understanding and reasoning abilities, and a curated benchmark for Echocardiology, ECBench, evaluating generalizability to unfamiliar specialty. The values marked in \reproduce{gray} indicate the results reproduced with the officially released checkpoint. Accuracy is used as the evaluation metric. ``-" denotes the checkpoint is not available for testing. 
    }
    \label{tab:main}
    \centering
    
    \begin{adjustbox}{width=1.0\textwidth}
        \begin{tabular}{lcccccccc}
        \toprule
        \multirow{2}{*}{\textbf{Model}} & \multirow{2}{*}{\textbf{Parameter}} & \multicolumn{3}{c}{\textbf{MedXpertQA-MM}} & \multirow{2}{*}{\textbf{MMMU-H\&M}} & \multicolumn{2}{c}{\textbf{MMMU-Pro-H\&M}} & \multirow{2}{*}{\textbf{ECBench}}\\ 
        \cmidrule{3-5} \cmidrule{7-8}
        & & Reasoning & Understanding & Overall & & 4 options & 10 options \\
        \midrule
        \midrule
        \multicolumn{9}{c}{\textit{Agentic Search MLLM}} \\
        \midrule
        Visual-ARFT~\citep{liu2025visual2} & 7B & \reproduce{21.0} & \reproduce{22.2} & \reproduce{22.0} & \reproduce{58.6} & \reproduce{47.6} & \reproduce{30.8} & \reproduce{41.0} \\
        MMSearch-R1~\citep{wu2025mmsearch} & 7B & \reproduce{22.3} & \reproduce{23.3} & \reproduce{22.6} & \reproduce{56.6} & \reproduce{43.4} & \reproduce{26.9} & \reproduce{41.1} \\
        \midrule
        \multicolumn{9}{c}{\textit{Generalist Medical MLLM}} \\
        \midrule
        MedRegA~\citep{wang2024interpretable} & 40B & \reproduce{23.1} & \reproduce{28.3} & \reproduce{24.6} & \reproduce{47.6} & \reproduce{43.4} & \reproduce{25.2} & \reproduce{37.6} \\
        HuatuoGPT-Vision~\citep{chen2024huatuogpt-2} & 34B & \reproduce{20.2} & \reproduce{26.4} & \reproduce{21.9} & 54.4 & \reproduce{42.0} & \reproduce{31.5} & \reproduce{43.1} \\
        MedGemma~\citep{sellergren2025medgemma} & 4B & - & - & 24.4 & \reproduce{47.3} & \reproduce{43.7} & \reproduce{32.9} & \reproduce{41.5} \\
        Lingshu~\citep{xu2025lingshu} & 7B & - & - & 26.7 & 54.0 & \reproduce{50.0} & \reproduce{37.1} & \reproduce{44.6} \\
        \midrule
        \multicolumn{9}{c}{\textit{Reasoning Medical MLLM}} \\
        \midrule
        MedVLM-R1~\citep{pan2025medvlm} & 2B & \reproduce{20.3} & \reproduce{19.5} & \reproduce{20.1} & \reproduce{43.5} & \reproduce{28.3} & \reproduce{18.5} & \reproduce{26.2} \\
        Med-R1~\citep{lai2025med} & 2B & \reproduce{21.8} & \reproduce{20.8} & \reproduce{21.5} & \reproduce{42.7} & \reproduce{33.9} & \reproduce{23.8} & \reproduce{36.7} \\
        GMAI-VL-R1~\citep{su2025gmai} & 7B & - & - & 23.8 & 57.3 & - & 34.0 & - \\
        XReasoner-Med~\citep{liu2025x} & 7B & - & - & 25.9 & 63.5 & - & 40.0 & - \\
        MedE$^2$~\citep{mu2025elicit} & 7B & 25.8 & 28.5 & 26.5 & \underline{66.0} & - & 38.8 & - \\
        MedCCO~\citep{rui2025improving} & 7B & 23.2 & 23.6 & 23.3 & 59.3 & - & - & - \\
        Chiron-o1~\citep{sun2025enhancing} & 8B & 23.3 & 25.1 & 23.8 & 54.6 & \reproduce{36.7} & \reproduce{24.5} & \reproduce{36.9} \\
        \rowcolor{highlight!30}\textbf{\textsc{Med-RwR} (Ours)} & 7B & \underline{26.2} & \underline{29.6} & \underline{27.2} & 65.5 & \underline{52.5} & \underline{43.7} & \underline{51.1} \\
        \rowcolor{highlight!60}\textbf{\textsc{Med-RwR+CDIR} (Ours) }& 7B & \textbf{26.6} & \textbf{29.7} & \textbf{27.5} & \textbf{66.2} & \textbf{52.8} & \textbf{44.1} & \textbf{51.9} \\
        \bottomrule
        \end{tabular}
    \end{adjustbox}
\end{table}


\begin{table}[t]
    \caption{Ablation Studies for Training Stages. ``Text-only" denotes the first training stage with text-only data, and ``Multimodal" is the second stage extending to multimodal data. To ensure fair comparison, only accuracy and format rewards are applied for all implementations.}
    \label{tab:stages123}
    \centering
    \begin{adjustbox}{width=0.7 \textwidth}
        \begin{tabular}{cccccc}
        \toprule
        \multicolumn{2}{c}{\textbf{Training Stage}} & \multirow{2}{*}{\textbf{MedXpertQA-MM}} & \multirow{2}{*}{\textbf{MMMU-H\&M}} & \multirow{2}{*}{\textbf{MMMU-Pro-H\&M}} & \multirow{2}{*}{\textbf{ECBench}} \\
        \cmidrule{1-2} 
        Text-only & Multimodal \\
                \midrule
        \midrule
        \no & \no & 22.4 & 56.5 & 31.8 & 40.4 \\
        \yes & \no  & 22.1 & 61.3 & 34.3 & 42.2 \\
        \no & \yes  & 24.7 & 59.3 & 36.3 & 43.0  \\
        \rowcolor{highlight!30} \yes & \yes & 25.1 & 60.0 & 37.1 & 45.3 \\
        \bottomrule
        \end{tabular}
    \end{adjustbox}
    \vspace{-1em}
\end{table}

\begin{table}[t]
    \caption{Ablation Studies for Reward Design. ``Query" and ``Conf" indicate Query Semantic Reward and Confidence Gain Reward. Accuracy and format rewards are default reward configurations.}
    \label{tab:reward123}
    \centering
    \begin{adjustbox}{width=0.7\textwidth}
        \begin{tabular}{cccccccc}
        \toprule
        \multicolumn{2}{c}{\textbf{Reward Design}} & \multirow{2}{*}{\textbf{MedXpertQA-MM}} & \multirow{2}{*}{\textbf{MMMU-H\&M}} & \multirow{2}{*}{\textbf{MMMU-Pro-H\&M}} & \multirow{2}{*}{\textbf{ECBench}} \\
        \cmidrule{1-2} 
        {Query} & {Conf} \\
                \midrule
        \midrule
        \no & \no & 25.1 & 60.0 & 37.1 & 45.3 \\
        \yes & \no & 25.4 & 62.8 & 38.8 & 47.2 \\
        \no & \yes & 26.2 & 62.0 & 37.1 & 45.9 \\
         \rowcolor{highlight!30} \yes & \yes & \textbf{27.2} & \textbf{65.5} & \textbf{43.7} & \textbf{51.1} \\
        \bottomrule
        \end{tabular}
    \end{adjustbox}
\end{table}

We report accuracy as the evaluation metric across all experiments. As shown in Table~\ref{tab:main}, our model achieves the highest accuracy on three complex reasoning benchmarks, showing competitive performance to larger general-purpose models despite significantly fewer parameters. 
\modelname{} surpasses the State-of-the-Art reasoning medical MLLM by 1\% (27.5\% \versus \ 26.5\%) on MedXpertQA-MM and 4.1\% (44.1\% \versus \ 40.0\%) on MMMU-Pro-H\&M, demonstrating the efficacy of external knowledge retrieval beyond solely reasoning with internal knowledge.
In addition to public benchmarks, we also perform experiments on the specialized benchmark for echocardiography, ECBench. Echocardiography is particularly challenging due to data scarcity and the high demand for clinical expertise, comprising less than 2\% of both the public training corpus and ours (see Figure~\ref{fig:data-distribution}). 
Despite not being specifically trained on domain-specific echocardiographic data, our model achieves an improved accuracy of 51.9\%. This highlights the generalizability of our model to specialized medical domains by actively incorporating the external domain-specific knowledge during reasoning.
Compared to Agentic Search MLLMs, our model's superior performance can be attributed to its ability to formulate more effective and context-aware search queries tailored to medical scenarios.

        
\vspace{-0.5em}
\subsection{Ablation Study}
\vspace{-0.5em}
\subsubsection{Training Design Analysis}

\textbf{Two-Stage Training.} We validate the necessity of two-stage training by comparing it with text-only training and direct multimodal training strategies. 
Table~\ref{tab:stages123} demonstrate that both alternatives can improve the reasoning-with-retrieval capability and enhance model performance to some extent. However, employing RL training directly with multimodal data yields substantially smaller benefits than pretraining with text-only data first. We attribute this to the potential of text-only data to unleash the language model's reasoning ability, which manifests as more comprehensive explanations after training. In contrast, direct multimodal training requires the model to simultaneously learn multimodal alignment and reasoning-with-retrieval abilities, which may amplify complexity and hinder the development of analytical skills. 

\textbf{Reward Design.} To evaluate the impact of our proposed rewards for retrieval ability incentivization,  we incrementally add them into the base model trained on the basic format and accuracy rewards. Table~\ref{tab:reward123} show that both rewards can improve the performance, as evidenced by the increased accuracy. This suggests that the query semantic reward drives the model to actively seek relevant documents aligned with multimodal inputs, while the confidence gain reward ensures that the retrieved knowledge genuinely benefits final decision making.

\begin{figure*}[t]
	\centering
	\includegraphics[width=1.0\textwidth]{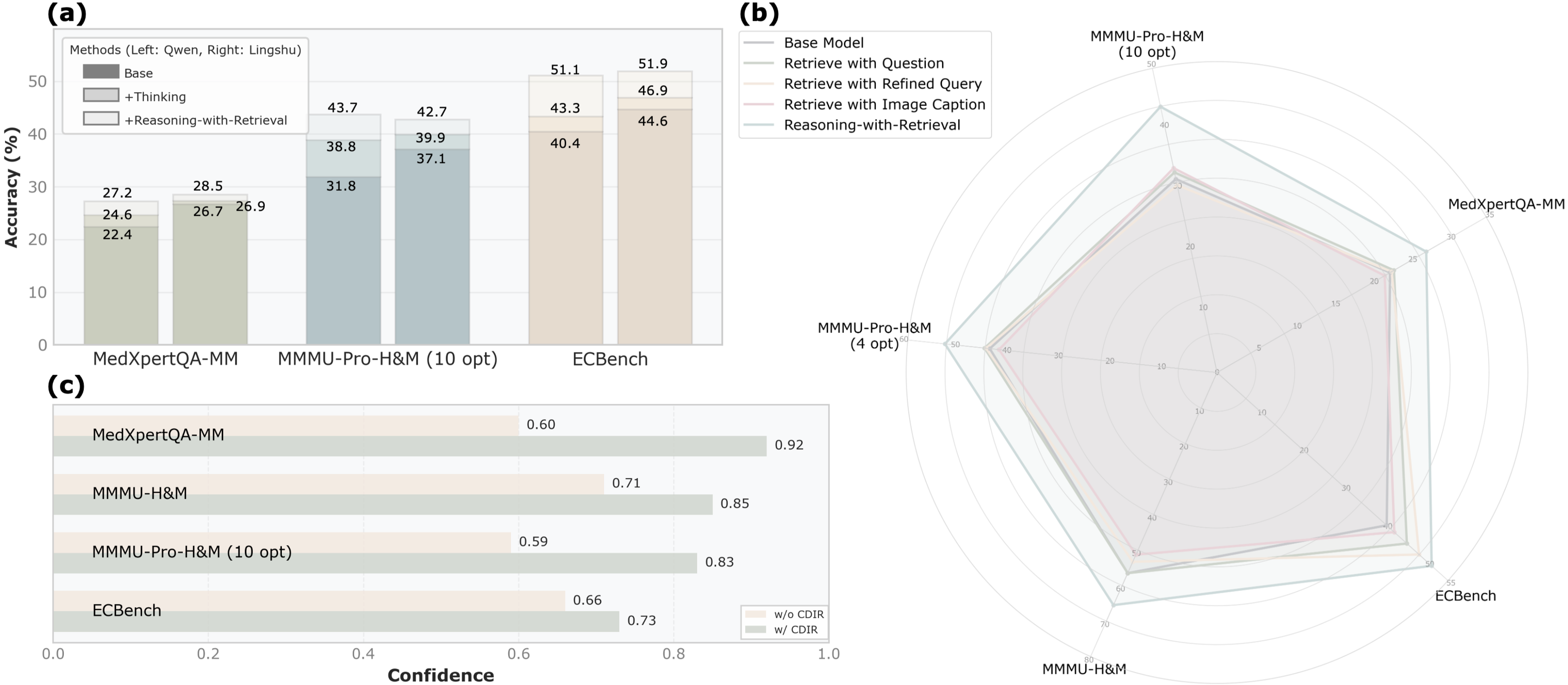}
	\caption{(a) Ablation Study of the Reasoning-with-Retrieval Capability: comparing the vanilla model, the reasoning model, and the reasoning-with-retrieval model on two architectures, QwenVL-2.5-7B (left) and Lingshu-7B (right). (b) Comparison with Training-Free RAG methods, which apply the original question, MLLM-reformulated question and MLLM-generated caption as the query, respectively. 
    (c) Confidence Gain after Confidence-Driven Image Re-retrieval.
    \vspace{-1em}
    }
	\label{fig:exp}
\end{figure*}

\textbf{Reasoning-with-Retrieval Capability.} 
To assess the effectiveness of the reasoning-with-retrieval framework, 
we conduct a comparative analysis against two baselines: the base model and the reasoning model trained with reinforcement learning, which only promotes reasoning ability but excludes query and retrieval actions. We apply two base architectures for ablation, QwenVL-2.5-7B~\citep{bai2025qwen2} and Lingshu-7B~\citep{xu2025lingshu}. Figure~\ref{fig:exp} (a) demonstrates that incorporating reasoning-with-retrieval capabilities improves performance regardless of whether the base model have undergone medical pretraining. This improvement stems from the model's enhanced ability to actively seek relevant information from external knowledge sources during the reasoning process.

\vspace{-0.5em}
\subsubsection{Knowledge Retrieval Analysis}
\textbf{Comparison against Training-Free RAG Method with Different Query Formulation.} To demonstrate how our proposed RL training strategy enhances medical MLLM's interaction with the external medical knowledge base, we compare \modelname{} with training-free RAG method, where the query is formulated from the input question to retrieve information during inference. We implement three query formulation approaches: (1) Retrieve with Question: original question as the query; (2) Retrieve with Refined Query: MLLM-refined question as the query; (3) Retrieve with Image Caption: MLLM-generated image caption as the query. Results are shown in Figure~\ref{fig:exp} (b). While RAG methods achieve moderate improvements over the base model, our framework demonstrates more effective engagement with the medical knowledge base. This is due to the tailored reward design, which elicits targeted retrieval and integration of essential knowledge.

\textbf{Confidence-Driven Image Re-retrieval.} Confidence-Driven Image Re-retrieval (\textsc{CDIR}) selectively triggers image re-retrieval when the model's initial confidence falls below a predefined threshold of 0.8. As demonstrated in Table~\ref{tab:main}, \textsc{CDIR} provides consistent performance improvements across all the benchmarks while maintaining a lightweight inference overhead.
Moreover, Figure~\ref{fig:exp} presents the average confidence of the uncertain cases. \textsc{CDIR} substantially improves model confidence by 0.12 to 0.21 across datasets, enabling more accurate and more reliable decision-making.
Supplementary experiments and case studies are included in Appendix~\ref{sec:supl-exp} and~\ref{sec:case}.

\vspace{-0.5em}
\section{Conclusion}
\vspace{-0.5em}
In this paper, we introduce \modelname{}, the first comprehensive Multimodal Medical Reasoning-with-Retrieval framework. We propose a two-stage reinforcement learning strategy guided by rewards, facilitating proactive retrieval of knowledge aligned with both visual findings and textual contexts. To augment potentially insufficient retrieval, we design a confidence-driven image re-retrieval mechanism for test-time scaling when model uncertainty is detected. Experiments show that \modelname{} achieves significant improvements on public benchmarks, demonstrating the crucial role of self-initiated external knowledge retrieval in complex medical reasoning.
To evaluate generalizability, we evaluate on a specialized echocardiography benchmark, ECBench, which highlights the advantage of our model in leveraging external knowledge to adapt to unfamiliar domains. 

\bibliography{iclr2026_conference}

\begin{thebibliography}{67}
\providecommand{\natexlab}[1]{#1}
\providecommand{\url}[1]{\texttt{#1}}
\expandafter\ifx\csname urlstyle\endcsname\relax
  \providecommand{\doi}[1]{doi: #1}\else
  \providecommand{\doi}{doi: \begingroup \urlstyle{rm}\Url}\fi

\bibitem[Asai et~al.(2024)Asai, Wu, Wang, Sil, and Hajishirzi]{asai2024self}
Akari Asai, Zeqiu Wu, Yizhong Wang, Avirup Sil, and Hannaneh Hajishirzi.
\newblock Self-rag: Learning to retrieve, generate, and critique through self-reflection.
\newblock 2024.

\bibitem[Bai et~al.(2025)Bai, Chen, Liu, Wang, Ge, Song, Dang, Wang, Wang, Tang, et~al.]{bai2025qwen2}
Shuai Bai, Keqin Chen, Xuejing Liu, Jialin Wang, Wenbin Ge, Sibo Song, Kai Dang, Peng Wang, Shijie Wang, Jun Tang, et~al.
\newblock Qwen2.5-vl technical report.
\newblock \emph{arXiv preprint arXiv:2502.13923}, 2025.

\bibitem[Bannur et~al.(2024)Bannur, Bouzid, Castro, Schwaighofer, Thieme, Bond-Taylor, Ilse, P{\'e}rez-Garc{\'\i}a, Salvatelli, Sharma, et~al.]{bannur2024maira}
Shruthi Bannur, Kenza Bouzid, Daniel~C Castro, Anton Schwaighofer, Anja Thieme, Sam Bond-Taylor, Maximilian Ilse, Fernando P{\'e}rez-Garc{\'\i}a, Valentina Salvatelli, Harshita Sharma, et~al.
\newblock Maira-2: Grounded radiology report generation.
\newblock \emph{arXiv preprint arXiv:2406.04449}, 2024.

\bibitem[Chen et~al.(2024{\natexlab{a}})Chen, Xiao, Zhang, Luo, Lian, and Liu]{chen2024m3}
Jianlyu Chen, Shitao Xiao, Peitian Zhang, Kun Luo, Defu Lian, and Zheng Liu.
\newblock M3-embedding: Multi-linguality, multi-functionality, multi-granularity text embeddings through self-knowledge distillation.
\newblock In \emph{Findings of the Association for Computational Linguistics ACL 2024}, pp.\  2318--2335, 2024{\natexlab{a}}.

\bibitem[Chen et~al.(2024{\natexlab{b}})Chen, Cai, Ji, Wang, Liu, Wang, Hou, and Wang]{chen2024huatuogpt}
Junying Chen, Zhenyang Cai, Ke~Ji, Xidong Wang, Wanlong Liu, Rongsheng Wang, Jianye Hou, and Benyou Wang.
\newblock Huatuogpt-o1, towards medical complex reasoning with llms.
\newblock \emph{arXiv preprint arXiv:2412.18925}, 2024{\natexlab{b}}.

\bibitem[Chen et~al.(2024{\natexlab{c}})Chen, Gui, Ouyang, Gao, Chen, Chen, Wang, Zhang, Cai, Ji, et~al.]{chen2024huatuogpt-2}
Junying Chen, Chi Gui, Ruyi Ouyang, Anningzhe Gao, Shunian Chen, Guiming~Hardy Chen, Xidong Wang, Ruifei Zhang, Zhenyang Cai, Ke~Ji, et~al.
\newblock Huatuogpt-vision, towards injecting medical visual knowledge into multimodal llms at scale.
\newblock \emph{arXiv preprint arXiv:2406.19280}, 2024{\natexlab{c}}.

\bibitem[Ding et~al.(2025)Ding, Huang, Fang, Liao, Jiang, Li, Zhao, and Wang]{ding2025promed}
Hongxin Ding, Baixiang Huang, Yue Fang, Weibin Liao, Xinke Jiang, Zheng Li, Junfeng Zhao, and Yasha Wang.
\newblock Promed: Shapley information gain guided reinforcement learning for proactive medical llms.
\newblock \emph{arXiv preprint arXiv:2508.13514}, 2025.

\bibitem[Ding et~al.(2024)Ding, Chu, Pi, Wang, and Li]{ding2024hia}
Xinpeng Ding, Yongqiang Chu, Renjie Pi, Hualiang Wang, and Xiaomeng Li.
\newblock Hia: Towards chinese multimodal llms for comparative high-resolution joint diagnosis.
\newblock In \emph{International Conference on Medical Image Computing and Computer-Assisted Intervention}, pp.\  575--586. Springer, 2024.

\bibitem[Dong et~al.(2025)Dong, Liu, Sun, Yang, Hu, Rao, and Liu]{dong2025insight}
Yuhao Dong, Zuyan Liu, Hai-Long Sun, Jingkang Yang, Winston Hu, Yongming Rao, and Ziwei Liu.
\newblock Insight-v: Exploring long-chain visual reasoning with multimodal large language models.
\newblock In \emph{Proceedings of the Computer Vision and Pattern Recognition Conference}, pp.\  9062--9072, 2025.

\bibitem[Du et~al.(2025)Du, Liu, Li, Zhao, Huo, Wang, Chen, Liu, Wang, and Wen]{du2025virgo}
Yifan Du, Zikang Liu, Yifan Li, Wayne~Xin Zhao, Yuqi Huo, Bingning Wang, Weipeng Chen, Zheng Liu, Zhongyuan Wang, and Ji-Rong Wen.
\newblock Virgo: A preliminary exploration on reproducing o1-like mllm.
\newblock \emph{arXiv preprint arXiv:2501.01904}, 2025.

\bibitem[Guo et~al.(2025)Guo, Yang, Zhang, Song, Zhang, Xu, Zhu, Ma, Wang, Bi, et~al.]{guo2025deepseek}
Daya Guo, Dejian Yang, Haowei Zhang, Junxiao Song, Ruoyu Zhang, Runxin Xu, Qihao Zhu, Shirong Ma, Peiyi Wang, Xiao Bi, et~al.
\newblock Deepseek-r1: Incentivizing reasoning capability in llms via reinforcement learning.
\newblock \emph{Nature}, 645:\penalty0 633--638, 2025.

\bibitem[He et~al.(2020)He, Cai, Wei, Zhang, Mou, Xing, and Xie]{he2020pathological}
Xuehai He, Zhuo Cai, Wenlan Wei, Yichen Zhang, Luntian Mou, Eric Xing, and Pengtao Xie.
\newblock Pathological visual question answering.
\newblock \emph{arXiv preprint arXiv:2010.12435}, 2020.

\bibitem[Hu et~al.(2024)Hu, Li, Lu, Shao, He, Qiao, and Luo]{hu2024omnimedvqa}
Yutao Hu, Tianbin Li, Quanfeng Lu, Wenqi Shao, Junjun He, Yu~Qiao, and Ping Luo.
\newblock Omnimedvqa: A new large-scale comprehensive evaluation benchmark for medical lvlm.
\newblock In \emph{Proceedings of the IEEE/CVF Conference on Computer Vision and Pattern Recognition}, pp.\  22170--22183, 2024.

\bibitem[Huang et~al.(2025{\natexlab{a}})Huang, Jia, Zhai, Cao, Ye, Zhao, Xu, Hu, and Lin]{huang2025vision}
Wenxuan Huang, Bohan Jia, Zijie Zhai, Shaosheng Cao, Zheyu Ye, Fei Zhao, Zhe Xu, Yao Hu, and Shaohui Lin.
\newblock Vision-r1: Incentivizing reasoning capability in multimodal large language models.
\newblock \emph{arXiv preprint arXiv:2503.06749}, 2025{\natexlab{a}}.

\bibitem[Huang et~al.(2025{\natexlab{b}})Huang, Geng, Hua, Huang, Zou, Zhang, Liu, and Zhang]{huang2025o1}
Zhongzhen Huang, Gui Geng, Shengyi Hua, Zhen Huang, Haoyang Zou, Shaoting Zhang, Pengfei Liu, and Xiaofan Zhang.
\newblock O1 replication journey--part 3: Inference-time scaling for medical reasoning.
\newblock \emph{arXiv preprint arXiv:2501.06458}, 2025{\natexlab{b}}.

\bibitem[Jeong et~al.(2024)Jeong, Sohn, Sung, and Kang]{jeong2024improving}
Minbyul Jeong, Jiwoong Sohn, Mujeen Sung, and Jaewoo Kang.
\newblock Improving medical reasoning through retrieval and self-reflection with retrieval-augmented large language models.
\newblock \emph{Bioinformatics}, 40\penalty0 (Supplement\_1):\penalty0 i119--i129, 2024.

\bibitem[Jin et~al.(2025)Jin, Zeng, Yue, Yoon, Arik, Wang, Zamani, and Han]{jin2025search}
Bowen Jin, Hansi Zeng, Zhenrui Yue, Jinsung Yoon, Sercan~O Arik, Dong Wang, Hamed Zamani, and Jiawei Han.
\newblock Search-r1: Training {LLM}s to reason and leverage search engines with reinforcement learning.
\newblock In \emph{Second Conference on Language Modeling}, 2025.
\newblock URL \url{https://openreview.net/forum?id=Rwhi91ideu}.

\bibitem[Jin et~al.(2021)Jin, Pan, Oufattole, Weng, Fang, and Szolovits]{jin2021disease}
Di~Jin, Eileen Pan, Nassim Oufattole, Wei-Hung Weng, Hanyi Fang, and Peter Szolovits.
\newblock What disease does this patient have? a large-scale open domain question answering dataset from medical exams.
\newblock \emph{Applied Sciences}, 11\penalty0 (14):\penalty0 6421, 2021.

\bibitem[Jin et~al.(2023)Jin, Kim, Chen, Comeau, Yeganova, Wilbur, and Lu]{jin2023medcpt}
Qiao Jin, Won Kim, Qingyu Chen, Donald~C Comeau, Lana Yeganova, W~John Wilbur, and Zhiyong Lu.
\newblock Medcpt: Contrastive pre-trained transformers with large-scale pubmed search logs for zero-shot biomedical information retrieval.
\newblock \emph{Bioinformatics}, 39\penalty0 (11):\penalty0 btad651, 2023.

\bibitem[Kraljevic et~al.(2021)Kraljevic, Searle, Shek, Roguski, Noor, Bean, Mascio, Zhu, Folarin, Roberts, Bendayan, Richardson, Stewart, Shah, Wong, Ibrahim, Teo, and Dobson]{Kraljevic2021-ln}
Zeljko Kraljevic, Thomas Searle, Anthony Shek, Lukasz Roguski, Kawsar Noor, Daniel Bean, Aurelie Mascio, Leilei Zhu, Amos~A Folarin, Angus Roberts, Rebecca Bendayan, Mark~P Richardson, Robert Stewart, Anoop~D Shah, Wai~Keong Wong, Zina Ibrahim, James~T Teo, and Richard J~B Dobson.
\newblock Multi-domain clinical natural language processing with {MedCAT}: The medical concept annotation toolkit.
\newblock \emph{Artif. Intell. Med.}, 117:\penalty0 102083, July 2021.
\newblock ISSN 0933-3657.
\newblock \doi{10.1016/j.artmed.2021.102083}.

\bibitem[Lai et~al.(2025)Lai, Zhong, Li, Zhao, and Yang]{lai2025med}
Yuxiang Lai, Jike Zhong, Ming Li, Shitian Zhao, and Xiaofeng Yang.
\newblock Med-r1: Reinforcement learning for generalizable medical reasoning in vision-language models.
\newblock \emph{arXiv preprint arXiv:2503.13939}, 2025.

\bibitem[Lau et~al.(2018)Lau, Gayen, Ben~Abacha, and Demner-Fushman]{lau2018dataset}
Jason~J Lau, Soumya Gayen, Asma Ben~Abacha, and Dina Demner-Fushman.
\newblock A dataset of clinically generated visual questions and answers about radiology images.
\newblock \emph{Scientific data}, 5\penalty0 (1):\penalty0 1--10, 2018.

\bibitem[Liu et~al.(2021)Liu, Zhan, Xu, Ma, Yang, and Wu]{liu2021slake}
Bo~Liu, Li-Ming Zhan, Li~Xu, Lin Ma, Yan Yang, and Xiao-Ming Wu.
\newblock Slake: A semantically-labeled knowledge-enhanced dataset for medical visual question answering.
\newblock In \emph{2021 IEEE 18th International Symposium on Biomedical Imaging (ISBI)}, pp.\  1650--1654. IEEE, 2021.

\bibitem[Liu et~al.(2025{\natexlab{a}})Liu, Zhang, Qin, Ossowski, Gu, Jin, Kiblawi, Preston, Wei, Vozila, et~al.]{liu2025x}
Qianchu Liu, Sheng Zhang, Guanghui Qin, Timothy Ossowski, Yu~Gu, Ying Jin, Sid Kiblawi, Sam Preston, Mu~Wei, Paul Vozila, et~al.
\newblock X-reasoner: Towards generalizable reasoning across modalities and domains.
\newblock \emph{arXiv preprint arXiv:2505.03981}, 2025{\natexlab{a}}.

\bibitem[Liu et~al.(2025{\natexlab{b}})Liu, Sun, Zang, Dong, Cao, Duan, Lin, and Wang]{liu2025visual}
Ziyu Liu, Zeyi Sun, Yuhang Zang, Xiaoyi Dong, Yuhang Cao, Haodong Duan, Dahua Lin, and Jiaqi Wang.
\newblock Visual-rft: Visual reinforcement fine-tuning.
\newblock \emph{arXiv preprint arXiv:2503.01785}, 2025{\natexlab{b}}.

\bibitem[Liu et~al.(2025{\natexlab{c}})Liu, Zang, Zou, Liang, Dong, Cao, Duan, Lin, and Wang]{liu2025visual2}
Ziyu Liu, Yuhang Zang, Yushan Zou, Zijian Liang, Xiaoyi Dong, Yuhang Cao, Haodong Duan, Dahua Lin, and Jiaqi Wang.
\newblock Visual agentic reinforcement fine-tuning.
\newblock \emph{arXiv preprint arXiv:2505.14246}, 2025{\natexlab{c}}.

\bibitem[Lu et~al.(2024)Lu, Chen, Williamson, Chen, Zhao, Chow, Ikemura, Kim, Pouli, Patel, et~al.]{lu2024multimodal}
Ming~Y Lu, Bowen Chen, Drew~FK Williamson, Richard~J Chen, Melissa Zhao, Aaron~K Chow, Kenji Ikemura, Ahrong Kim, Dimitra Pouli, Ankush Patel, et~al.
\newblock A multimodal generative ai copilot for human pathology.
\newblock \emph{Nature}, 634\penalty0 (8033):\penalty0 466--473, 2024.

\bibitem[Meng et~al.(2025)Meng, Du, Liu, Zhou, Lu, Fu, Han, Shi, Wang, He, et~al.]{meng2025mm}
Fanqing Meng, Lingxiao Du, Zongkai Liu, Zhixiang Zhou, Quanfeng Lu, Daocheng Fu, Tiancheng Han, Botian Shi, Wenhai Wang, Junjun He, et~al.
\newblock Mm-eureka: Exploring the frontiers of multimodal reasoning with rule-based reinforcement learning.
\newblock \emph{arXiv preprint arXiv:2503.07365}, 2025.

\bibitem[Mu et~al.(2025)Mu, Huang, Zhu, Zhao, Zhang, and Zhang]{mu2025elicit}
Linjie Mu, Zhongzhen Huang, Yakun Zhu, Xiangyu Zhao, Shaoting Zhang, and Xiaofan Zhang.
\newblock Elicit and enhance: Advancing multimodal reasoning in medical scenarios.
\newblock \emph{arXiv preprint arXiv:2505.23118}, 2025.

\bibitem[Ong et~al.(2024)Ong, Obey, Zheng, Cohan, and Schneider]{ong2024surgeryllm}
Chin~Siang Ong, Nicholas~T Obey, Yanan Zheng, Arman Cohan, and Eric~B Schneider.
\newblock Surgeryllm: a retrieval-augmented generation large language model framework for surgical decision support and workflow enhancement.
\newblock \emph{npj Digital Medicine}, 7\penalty0 (1):\penalty0 364, 2024.

\bibitem[Pai \& Varadarajan(2024)Pai and Varadarajan]{pai2024echocardiography}
Ramdas~G Pai and Padmini Varadarajan.
\newblock \emph{Echocardiography Board Review: 600 Multiple Choice Questions with Discussion}.
\newblock John Wiley \& Sons, 2024.

\bibitem[Pal et~al.(2022)Pal, Umapathi, and Sankarasubbu]{pal2022medmcqa}
Ankit Pal, Logesh~Kumar Umapathi, and Malaikannan Sankarasubbu.
\newblock Medmcqa: A large-scale multi-subject multi-choice dataset for medical domain question answering.
\newblock In \emph{Conference on health, inference, and learning}, pp.\  248--260. PMLR, 2022.

\bibitem[Pan et~al.(2025)Pan, Liu, Wu, Liu, Zhu, Li, Chen, Ouyang, and Rueckert]{pan2025medvlm}
Jiazhen Pan, Che Liu, Junde Wu, Fenglin Liu, Jiayuan Zhu, Hongwei~Bran Li, Chen Chen, Cheng Ouyang, and Daniel Rueckert.
\newblock Medvlm-r1: Incentivizing medical reasoning capability of vision-language models (vlms) via reinforcement learning.
\newblock \emph{arXiv preprint arXiv:2502.19634}, 2025.

\bibitem[Peng et~al.(2025)Peng, Zhang, Zhang, You, Liu, Zhu, Yang, Xu, Geng, and Yang]{peng2025lmmr1empowering3blmms}
Yingzhe Peng, Gongrui Zhang, Miaosen Zhang, Zhiyuan You, Jie Liu, Qipeng Zhu, Kai Yang, Xingzhong Xu, Xin Geng, and Xu~Yang.
\newblock Lmm-r1: Empowering 3b lmms with strong reasoning abilities through two-stage rule-based rl, 2025.
\newblock URL \url{https://arxiv.org/abs/2503.07536}.

\bibitem[Qin et~al.(2025)Qin, Gamage~Nanayakkara, and Li]{qin2025multiagent}
Yi~Qin, Dinusara~Sasindu Gamage~Nanayakkara, and Xiaomeng Li.
\newblock { Multi-Agent Collaboration for Integrating Echocardiography Expertise in Multi-Modal Large Language Models }.
\newblock In \emph{proceedings of Medical Image Computing and Computer Assisted Intervention -- MICCAI 2025}, volume LNCS 15966. Springer Nature Switzerland, September 2025.

\bibitem[Rui et~al.(2025)Rui, Chen, Ma, and Wang]{rui2025improving}
Shaohao Rui, Kaitao Chen, Weijie Ma, and Xiaosong Wang.
\newblock Improving medical reasoning with curriculum-aware reinforcement learning.
\newblock \emph{arXiv preprint arXiv:2505.19213}, 2025.

\bibitem[Sellergren et~al.(2025)Sellergren, Kazemzadeh, Jaroensri, Kiraly, Traverse, Kohlberger, Xu, Jamil, Hughes, Lau, et~al.]{sellergren2025medgemma}
Andrew Sellergren, Sahar Kazemzadeh, Tiam Jaroensri, Atilla Kiraly, Madeleine Traverse, Timo Kohlberger, Shawn Xu, Fayaz Jamil, C{\'\i}an Hughes, Charles Lau, et~al.
\newblock Medgemma technical report.
\newblock \emph{arXiv preprint arXiv:2507.05201}, 2025.

\bibitem[Siegel et~al.(2014)Siegel, Gurudevan, Shiota, Tolstrup, Beigel, and Wunderlich]{siegel2014complex}
R.J. Siegel, S.V. Gurudevan, T.~Shiota, K.~Tolstrup, R.~Beigel, and N.~Wunderlich.
\newblock \emph{Complex Cases in Echocardiography}.
\newblock Wolters Kluwer Health/Lippincott Williams \& Wilkins, 2014.
\newblock ISBN 9781451176469.

\bibitem[Song et~al.(2025)Song, Jiang, Tian, Chen, Wu, Zhao, Min, Zhao, Fang, and Wen]{song2025r1}
Huatong Song, Jinhao Jiang, Wenqing Tian, Zhipeng Chen, Yuhuan Wu, Jiahao Zhao, Yingqian Min, Wayne~Xin Zhao, Lei Fang, and Ji-Rong Wen.
\newblock R1-searcher++: Incentivizing the dynamic knowledge acquisition of llms via reinforcement learning.
\newblock \emph{arXiv preprint arXiv:2505.17005}, 2025.

\bibitem[Sreedharan et~al.(2024)Sreedharan, Khanna, Moghekar, Dugar, and Collier]{sreedharan2024critical}
Roshni Sreedharan, Sandeep Khanna, Ajit Moghekar, Siddharth Dugar, and Patrick Collier.
\newblock \emph{Critical Care Echocardiography: A Self-Assessment Book}.
\newblock Springer Nature, 2024.

\bibitem[Su et~al.(2025)Su, Li, Liu, Ma, Ning, Tang, Ju, Ye, Chen, Hu, et~al.]{su2025gmai}
Yanzhou Su, Tianbin Li, Jiyao Liu, Chenglong Ma, Junzhi Ning, Cheng Tang, Sibo Ju, Jin Ye, Pengcheng Chen, Ming Hu, et~al.
\newblock Gmai-vl-r1: Harnessing reinforcement learning for multimodal medical reasoning.
\newblock \emph{arXiv preprint arXiv:2504.01886}, 2025.

\bibitem[Sun et~al.(2025{\natexlab{a}})Sun, Qiao, Guo, Fan, Hou, Jiang, Xie, Zhang, Huang, and Zhou]{sun2025zerosearch}
Hao Sun, Zile Qiao, Jiayan Guo, Xuanbo Fan, Yingyan Hou, Yong Jiang, Pengjun Xie, Yan Zhang, Fei Huang, and Jingren Zhou.
\newblock Zerosearch: Incentivize the search capability of llms without searching.
\newblock \emph{arXiv preprint arXiv:2505.04588}, 2025{\natexlab{a}}.

\bibitem[Sun et~al.(2025{\natexlab{b}})Sun, Jiang, Lou, Zhang, Li, Wang, Liu, Liu, and Wang]{sun2025enhancing}
Haoran Sun, Yankai Jiang, Wenjie Lou, Yujie Zhang, Wenjie Li, Lilong Wang, Mianxin Liu, Lei Liu, and Xiaosong Wang.
\newblock Enhancing step-by-step and verifiable medical reasoning in mllms.
\newblock \emph{arXiv preprint arXiv:2506.16962}, 2025{\natexlab{b}}.

\bibitem[Wang et~al.(2024)Wang, Wang, Yang, Mao, Yang, Shen, and Li]{wang2024interpretable}
Lehan Wang, Haonan Wang, Honglong Yang, Jiaji Mao, Zehong Yang, Jun Shen, and Xiaomeng Li.
\newblock Interpretable bilingual multimodal large language model for diverse biomedical tasks.
\newblock \emph{arXiv preprint arXiv:2410.18387}, 2024.

\bibitem[Wang et~al.(2025)Wang, Ding, Zeng, Chen, Chen, Wang, Xie, Huang, and Zhao]{wang2025vrag}
Qiuchen Wang, Ruixue Ding, Yu~Zeng, Zehui Chen, Lin Chen, Shihang Wang, Pengjun Xie, Fei Huang, and Feng Zhao.
\newblock Vrag-rl: Empower vision-perception-based rag for visually rich information understanding via iterative reasoning with reinforcement learning.
\newblock \emph{arXiv preprint arXiv:2505.22019}, 2025.

\bibitem[Wu et~al.(2023)Wu, Zhang, Zhang, Wang, and Xie]{wu2023towards}
Chaoyi Wu, Xiaoman Zhang, Ya~Zhang, Yanfeng Wang, and Weidi Xie.
\newblock Towards generalist foundation model for radiology by leveraging web-scale 2d\&3d medical data.
\newblock \emph{arXiv preprint arXiv:2308.02463}, 2023.

\bibitem[Wu et~al.(2025{\natexlab{a}})Wu, Deng, Li, Liu, You, Li, Ma, and Liu]{wu2025mmsearch}
Jinming Wu, Zihao Deng, Wei Li, Yiding Liu, Bo~You, Bo~Li, Zejun Ma, and Ziwei Liu.
\newblock Mmsearch-r1: Incentivizing lmms to search.
\newblock \emph{arXiv preprint arXiv:2506.20670}, 2025{\natexlab{a}}.

\bibitem[Wu et~al.(2025{\natexlab{b}})Wu, Deng, Li, Liu, Mi, Peng, Xu, Liu, Cho, Choi, et~al.]{wu2025medreason}
Juncheng Wu, Wenlong Deng, Xingxuan Li, Sheng Liu, Taomian Mi, Yifan Peng, Ziyang Xu, Yi~Liu, Hyunjin Cho, Chang-In Choi, et~al.
\newblock Medreason: Eliciting factual medical reasoning steps in llms via knowledge graphs.
\newblock \emph{arXiv preprint arXiv:2504.00993}, 2025{\natexlab{b}}.

\bibitem[Xia et~al.(2025)Xia, Zang, Gao, Li, and Zhou]{xia2025visionary}
Jiaer Xia, Yuhang Zang, Peng Gao, Yixuan Li, and Kaiyang Zhou.
\newblock Visionary-r1: Mitigating shortcuts in visual reasoning with reinforcement learning.
\newblock \emph{arXiv preprint arXiv:2505.14677}, 2025.

\bibitem[Xu et~al.(2025{\natexlab{a}})Xu, Nie, Wang, Chen, Li, Ning, Liu, Wang, Zhu, Liu, Li, and He]{xu2025medground}
Huihui Xu, Yuanpeng Nie, Hualiang Wang, Ying Chen, Wei Li, Junzhi Ning, Lihao Liu, Hongqiu Wang, Lei Zhu, Jiyao Liu, Xiaomeng Li, and Junjun He.
\newblock { MedGround-R1: Advancing Medical Image Grounding via Spatial-Semantic Rewarded Group Relative Policy Optimization }.
\newblock In \emph{proceedings of Medical Image Computing and Computer Assisted Intervention -- MICCAI 2025}, volume LNCS 15964. Springer Nature Switzerland, September 2025{\natexlab{a}}.

\bibitem[Xu et~al.(2025{\natexlab{b}})Xu, Chan, Li, Aljunied, Yuan, Wang, Xiao, Chen, Liu, Li, et~al.]{xu2025lingshu}
Weiwen Xu, Hou~Pong Chan, Long Li, Mahani Aljunied, Ruifeng Yuan, Jianyu Wang, Chenghao Xiao, Guizhen Chen, Chaoqun Liu, Zhaodonghui Li, et~al.
\newblock Lingshu: A generalist foundation model for unified multimodal medical understanding and reasoning.
\newblock \emph{arXiv preprint arXiv:2506.07044}, 2025{\natexlab{b}}.

\bibitem[Yang et~al.(2025{\natexlab{a}})Yang, Li, Yang, Zhang, Hui, Zheng, Yu, Gao, Huang, Lv, et~al.]{yang2025qwen3}
An~Yang, Anfeng Li, Baosong Yang, Beichen Zhang, Binyuan Hui, Bo~Zheng, Bowen Yu, Chang Gao, Chengen Huang, Chenxu Lv, et~al.
\newblock Qwen3 technical report.
\newblock \emph{arXiv preprint arXiv:2505.09388}, 2025{\natexlab{a}}.

\bibitem[Yang et~al.(2025{\natexlab{b}})Yang, Song, Qin, Wang, Wang, Ding, Zhang, Du, and Li]{yang2025multi}
Honglong Yang, Shanshan Song, Yi~Qin, Lehan Wang, Haonan Wang, Xinpeng Ding, Qixiang Zhang, Bodong Du, and Xiaomeng Li.
\newblock Multi-modal explainable medical ai assistant for trustworthy human-ai collaboration.
\newblock \emph{arXiv preprint arXiv:2505.06898}, 2025{\natexlab{b}}.

\bibitem[Yao et~al.(2024)Yao, Huang, Wu, Zhang, Wang, Liu, Wang, Song, Feng, Shen, et~al.]{yao2024mulberry}
Huanjin Yao, Jiaxing Huang, Wenhao Wu, Jingyi Zhang, Yibo Wang, Shunyu Liu, Yingjie Wang, Yuxin Song, Haocheng Feng, Li~Shen, et~al.
\newblock Mulberry: Empowering mllm with o1-like reasoning and reflection via collective monte carlo tree search.
\newblock \emph{arXiv preprint arXiv:2412.18319}, 2024.

\bibitem[Yu et~al.(2025{\natexlab{a}})Yu, Yao, Liu, Chen, Yin, Wang, Liao, Ye, Li, Yue, et~al.]{yu2025medreseacher}
Ailing Yu, Lan Yao, Jingnan Liu, Zhe Chen, Jiajun Yin, Yuan Wang, Xinhao Liao, Zhiling Ye, Ji~Li, Yun Yue, et~al.
\newblock Medreseacher-r1: Expert-level medical deep researcher via a knowledge-informed trajectory synthesis framework.
\newblock \emph{arXiv preprint arXiv:2508.14880}, 2025{\natexlab{a}}.

\bibitem[Yu et~al.(2025{\natexlab{b}})Yu, Zhang, Zhu, Yuan, Zuo, Yue, Fan, Liu, Liu, Liu, et~al.]{yu2025dapo}
Qiying Yu, Zheng Zhang, Ruofei Zhu, Yufeng Yuan, Xiaochen Zuo, Yu~Yue, Tiantian Fan, Gaohong Liu, Lingjun Liu, Xin Liu, et~al.
\newblock Dapo: An open-source llm reinforcement learning system at scale, 2025.
\newblock \emph{URL https://arxiv. org/abs/2503.14476}, 2025{\natexlab{b}}.

\bibitem[Yue et~al.(2024{\natexlab{a}})Yue, Ni, Zhang, Zheng, Liu, Zhang, Stevens, Jiang, Ren, Sun, et~al.]{yue2024mmmu}
Xiang Yue, Yuansheng Ni, Kai Zhang, Tianyu Zheng, Ruoqi Liu, Ge~Zhang, Samuel Stevens, Dongfu Jiang, Weiming Ren, Yuxuan Sun, et~al.
\newblock Mmmu: A massive multi-discipline multimodal understanding and reasoning benchmark for expert agi.
\newblock In \emph{Proceedings of the IEEE/CVF Conference on Computer Vision and Pattern Recognition}, pp.\  9556--9567, 2024{\natexlab{a}}.

\bibitem[Yue et~al.(2024{\natexlab{b}})Yue, Zheng, Ni, Wang, Zhang, Tong, Sun, Yu, Zhang, Sun, et~al.]{yue2024mmmu2}
Xiang Yue, Tianyu Zheng, Yuansheng Ni, Yubo Wang, Kai Zhang, Shengbang Tong, Yuxuan Sun, Botao Yu, Ge~Zhang, Huan Sun, et~al.
\newblock Mmmu-pro: A more robust multi-discipline multimodal understanding benchmark.
\newblock \emph{arXiv preprint arXiv:2409.02813}, 2024{\natexlab{b}}.

\bibitem[Zhang et~al.(2024)Zhang, Zhou, Adhikarla, Yan, Liu, Yu, Liu, Chen, Davison, Ren, et~al.]{zhang2024generalist}
Kai Zhang, Rong Zhou, Eashan Adhikarla, Zhiling Yan, Yixin Liu, Jun Yu, Zhengliang Liu, Xun Chen, Brian~D Davison, Hui Ren, et~al.
\newblock A generalist vision--language foundation model for diverse biomedical tasks.
\newblock \emph{Nature Medicine}, 30\penalty0 (11):\penalty0 3129--3141, 2024.

\bibitem[Zhang et~al.(2023{\natexlab{a}})Zhang, Xu, Usuyama, Xu, Bagga, Tinn, Preston, Rao, Wei, Valluri, et~al.]{zhang2023biomedclip}
Sheng Zhang, Yanbo Xu, Naoto Usuyama, Hanwen Xu, Jaspreet Bagga, Robert Tinn, Sam Preston, Rajesh Rao, Mu~Wei, Naveen Valluri, et~al.
\newblock Biomedclip: a multimodal biomedical foundation model pretrained from fifteen million scientific image-text pairs.
\newblock \emph{arXiv preprint arXiv:2303.00915}, 2023{\natexlab{a}}.

\bibitem[Zhang et~al.(2023{\natexlab{b}})Zhang, Wu, Zhao, Lin, Zhang, Wang, and Xie]{zhang2023pmc}
Xiaoman Zhang, Chaoyi Wu, Ziheng Zhao, Weixiong Lin, Ya~Zhang, Yanfeng Wang, and Weidi Xie.
\newblock Pmc-vqa: Visual instruction tuning for medical visual question answering.
\newblock \emph{arXiv preprint arXiv:2305.10415}, 2023{\natexlab{b}}.

\bibitem[Zhao et~al.(2025{\natexlab{a}})Zhao, Liu, Yang, and Miao]{zhao2025medrag}
Xuejiao Zhao, Siyan Liu, Su-Yin Yang, and Chunyan Miao.
\newblock Medrag: Enhancing retrieval-augmented generation with knowledge graph-elicited reasoning for healthcare copilot.
\newblock In \emph{Proceedings of the ACM on Web Conference 2025}, pp.\  4442--4457, 2025{\natexlab{a}}.

\bibitem[Zhao et~al.(2025{\natexlab{b}})Zhao, Huang, Hu, Wang, Mao, Zhang, Jiang, Wu, Ai, Wang, et~al.]{zhao2025swift}
Yuze Zhao, Jintao Huang, Jinghan Hu, Xingjun Wang, Yunlin Mao, Daoze Zhang, Zeyinzi Jiang, Zhikai Wu, Baole Ai, Ang Wang, et~al.
\newblock Swift: a scalable lightweight infrastructure for fine-tuning.
\newblock In \emph{Proceedings of the AAAI Conference on Artificial Intelligence}, volume~39, pp.\  29733--29735, 2025{\natexlab{b}}.

\bibitem[Zheng et~al.(2025{\natexlab{a}})Zheng, Sun, Wu, Zhao, Qiu, Yu, Sun, Wang, Zhang, and Xie]{zheng2025end}
Qiaoyu Zheng, Yuze Sun, Chaoyi Wu, Weike Zhao, Pengcheng Qiu, Yongguo Yu, Kun Sun, Yanfeng Wang, Ya~Zhang, and Weidi Xie.
\newblock End-to-end agentic rag system training for traceable diagnostic reasoning.
\newblock \emph{arXiv preprint arXiv:2508.15746}, 2025{\natexlab{a}}.

\bibitem[Zheng et~al.(2025{\natexlab{b}})Zheng, Fu, Hu, Cai, Ye, Lu, and Liu]{zheng2025deepresearcher}
Yuxiang Zheng, Dayuan Fu, Xiangkun Hu, Xiaojie Cai, Lyumanshan Ye, Pengrui Lu, and Pengfei Liu.
\newblock Deepresearcher: Scaling deep research via reinforcement learning in real-world environments.
\newblock \emph{arXiv preprint arXiv:2504.03160}, 2025{\natexlab{b}}.

\bibitem[Zhu et~al.(2025)Zhu, Wang, Chen, Liu, Ye, Gu, Tian, Duan, Su, Shao, et~al.]{zhu2025internvl3}
Jinguo Zhu, Weiyun Wang, Zhe Chen, Zhaoyang Liu, Shenglong Ye, Lixin Gu, Hao Tian, Yuchen Duan, Weijie Su, Jie Shao, et~al.
\newblock Internvl3: Exploring advanced training and test-time recipes for open-source multimodal models.
\newblock \emph{arXiv preprint arXiv:2504.10479}, 2025.

\bibitem[Zuo et~al.(2025)Zuo, Qu, Li, Chen, Zhu, Hua, Zhang, Ding, and Zhou]{zuo2025medxpertqa}
Yuxin Zuo, Shang Qu, Yifei Li, Zhangren Chen, Xuekai Zhu, Ermo Hua, Kaiyan Zhang, Ning Ding, and Bowen Zhou.
\newblock Medxpertqa: Benchmarking expert-level medical reasoning and understanding.
\newblock \emph{arXiv preprint arXiv:2501.18362}, 2025.

\end{thebibliography}
\bibliographystyle{iclr2026_conference}

\appendix
\newpage
{\LARGE\sc{Appendix\par}}
\section{Training Data Construction Details}
\label{sec:train-data-construct}
\subsection{Dataset Statistics}
\begin{figure*}[h]
	\centering
	\includegraphics[width=1.0\textwidth]{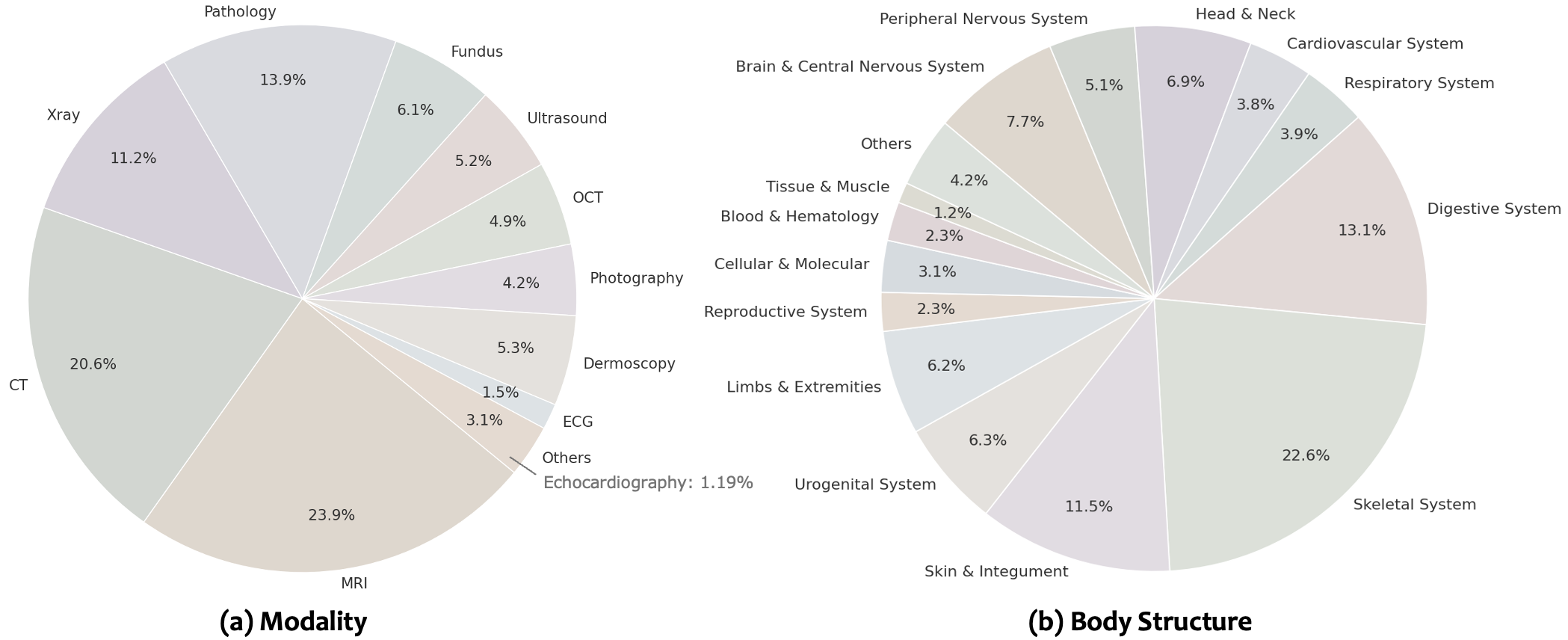}
	\caption{Modality and Body Structure Distribution of Our Constructed Multimodal Dataset.}
	\label{fig:data-distribution}
\end{figure*}

The Cluster-Reconstruct-Stratify data synthesis pipeline yields 6,500 multimodal question-answer pairs. We also verify that no instances in the training data overlap with evaluation benchmarks.
Figure~\ref{fig:data-distribution} presents the modality and body structure distribution of our constructed multimodal dataset. In statistics, our dataset covers at least 10 medical modalities of various body parts. Notably, echocardiography data comprises less than 2\% of our training corpus, reflecting severe underrepresentation.
\label{sec:data-stat}

\subsection{Prompts for Data Construction}
We curate training data from a public instruction tuning dataset, PubMedVision~\citep{chen2024huatuogpt-2}, following a two-step processing with GPT-4o.
For synthesizing visual question answering data, we first prompt GPT-4o to extract basic background, analysis and conclusion from the original data. The prompt is displayed in Table~\ref{tab:prompt-data-1}.

\begin{table}[h]
\caption{Prompt for Initial Information Extraction from Existing VQA Samples.}
\label{tab:prompt-data-1}
\centering
\begin{tabular}{p{0.9\linewidth}}
\toprule
\textbf{Prompt Template for Extracting Information from Original Data} \\
\midrule
\small{You will be provided with a visual question-answer pair, which includes an image, a question, and an answer. Your task is to analyze this information and extract the clinical information (patient basic information, medical history, or laboratory results), observation, analysis and conclusion from the given contents. Format the output as JSON dictionary.}
\\
\bottomrule
\end{tabular}
\end{table}

\begin{figure*}[h]
	\centering
	\includegraphics[width=0.95\textwidth]{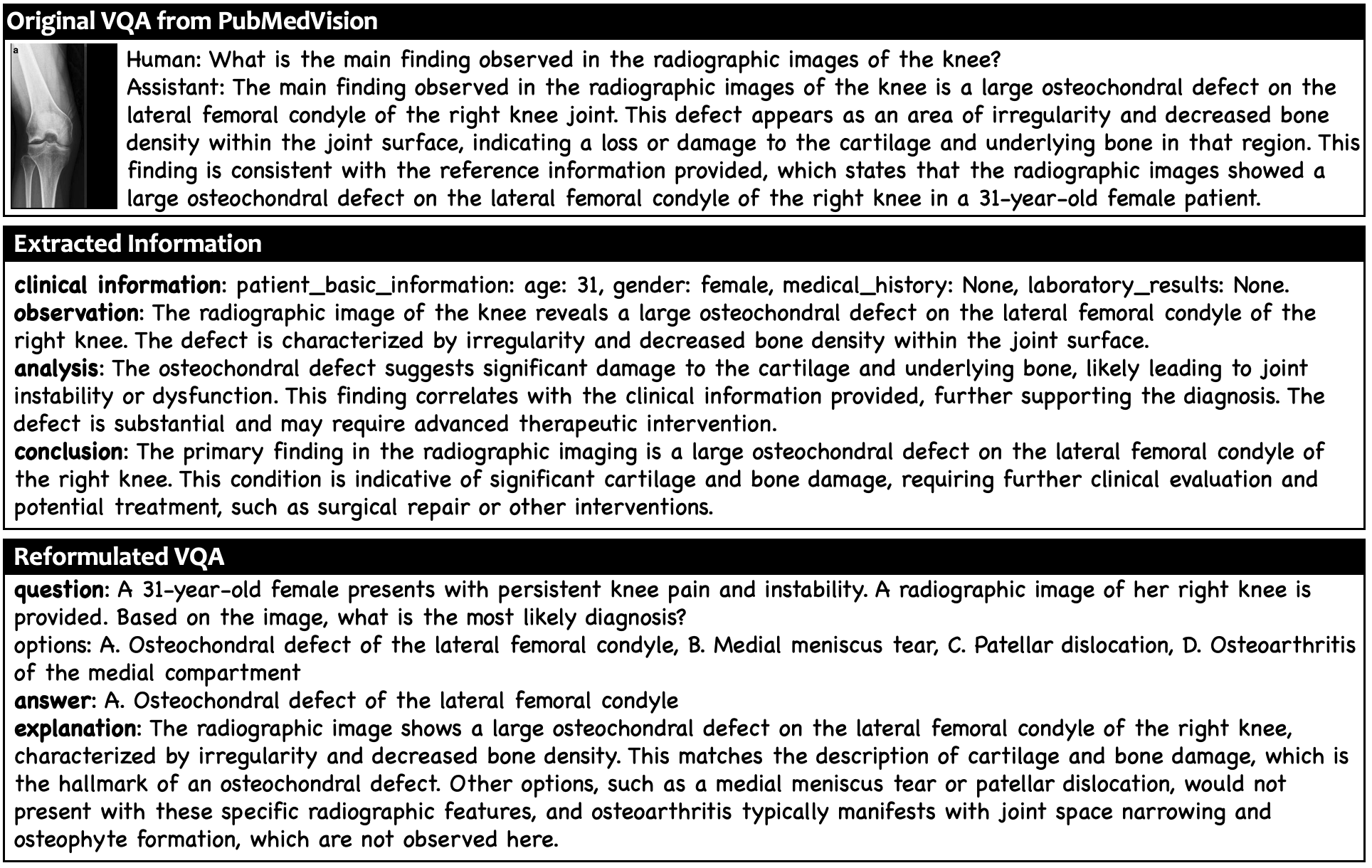}
	\caption{Illustration of the Data Synthesis Process.}
	\label{fig:train-sample}
\end{figure*}

Sequentially, we further prompt GPT-4o to formulate confusable options according to the provided descriptions, with the prompt in Table~\ref{tab:prompt-data-2}. In Figure~\ref{fig:train-sample}, we present an example to illustrate the complete data synthesis process.

\begin{table}[h]
\caption{Prompt for Training Data Construction Given Extracted Information.}
\label{tab:prompt-data-2}
\centering
\begin{tabular}{p{0.95\linewidth}}
\toprule
\textbf{Prompt Template for Constructing Training Data from the Extracted Information} \\
\midrule
\small{Please construct a standard and complex medical exam question and answer based on the example. You are provided with an image alongside with the observation, analysis and conclusion. Please follow the following instruction:}\\
\small{1. Create a new question in a complex and standard format. If clinical information about the patient is available, such as basic information, medical history, or laboratory results, incorporate these relevant clinical details into the question to provide appropriate context for the diagnostic scenario.}\\
\small{2. Since the original question requires an image, the constructed question should also require observing an image to answer. Do not indicate what the image contains.}\\
\small{3. Include the correct answer and up to three wrong but plausible/confusing options (no more than four options total). Randomize the position of the correct answer.}\\
\small{4. The constructed question should not contain any additional information that is not provided and mentioned in the given contexts. Only use the existing information.}\\
\small{Format the output as JSON containing:}\\
\small{``question": The question}\\
\small{``options": A list of answer choices, each prefixed with a letter label (e.g., ``A. ...", ``B. ...").}\\
\small{``answer": The correct answer, indicated by letter (e.g., ``B. ...").}\\
\small{``explanation": The explanation of the correct answer.}\\
\bottomrule
\end{tabular}
\end{table}

\section{EchoCardiography Benchmark (ECBench) Construction Details}
\label{sec:ecbench}
To establish a benchmark that provides objective evaluations for echocardiography, we utilized three authoritative echocardiography practice books as the foundational sources for the benchmark: (1) Critical Care Echocardiography: A Self-Assessment Book~\citep{sreedharan2024critical}, (2) Echocardiography Board Review: 600 Multiple Choice Questions with Discussion~\citep{pai2024echocardiography}, (3) Complex Cases in Echocardiography~\citep{siegel2014complex}. Importantly, these three practice books were not incorporated into the model development process to maintain fairness and prevent data leakage.

These books offer comprehensive coverage, including foundational knowledge of echocardiography and case studies relevant to echocardiographic diagnosis. Each question is formatted as a multiple-choice question, accompanied by its ground truth answer and an explanatory paragraph detailing the rationale. Furthermore, a significant proportion of the questions are supplemented with corresponding echocardiographic images for diagnostic observation.

Adopting the methodology outlined in~\cite{qin2025multiagent}, the construction of the ECBench benchmark involves a three-stage process: (1) extracting raw content from the practice books using the state-of-the-art Mistral OCR\footnote{\href{https://mistral.ai/news/mistral-ocr}{https://mistral.ai/news/mistral-ocr}}, which converts content into markdown format; (2) utilizing GPT-4o to structure the markdown content into a list of dictionaries, with each entry containing the question, its answer, the rationale, and any associated multimedia content; and (3) employing GPT-4o to refine the output by removing redundant inline references. The detailed prompts used for the extraction process are presented in Tables~\ref{tab:ecbench-1} and~\ref{tab:ecbench-2}.

\begin{table}[h]
\caption{Prompt for Organizing Raw Markdown Content.}
\label{tab:ecbench-1}
\centering
\begin{tabular}{p{0.9\linewidth}}
\toprule
\textbf{Prompt for Organizing Raw Markdown Content} \\
\midrule
\small{You are given a paragraph containing potential four types of content: questions, images in questions, answers, and explanations for answers. This paragraph is in markdown format. Your task is to organize the raw content into an structured dict, you should return a json list, which each of the item of the json dict looks like this: {`question':`(the complete question including the background information)', `candidates': `the candidate', `answer': `the ground truth answer', `explanation': `the explanation to the answer', `image': [The list of image paths that are involved in the question.]}. Note the image path should be only in the one in the markdown image reference format. Note that the stem (clinical background) of the question should also be properly included in each question, if included in the paragraph. If there's no content for a specific entry, fill the entry with 'None'. Do not summarize the content. Return the list of dict only. The paragraph is: [paragraph]}
\\
\bottomrule
\end{tabular}
\end{table}

\begin{table}[h]
\caption{Prompt for Clean Up Inline References.}
\label{tab:ecbench-2}
\centering
\begin{tabular}{p{0.9\linewidth}}
\toprule
\textbf{Prompt for Clean Up Inline References} \\
\midrule
\small{You are given a question. Please clean the content by: 1. remove inline image reference (e.g., Figure/Fig. 1.1 ..., Video/Vid. 1.1. ...), 2. if there's any candidate choice, remove them. Do not make other modifications. If there's no need for cleaning, just return the original question. Return the cleaned question only. This is the question: [question]}
\\
\bottomrule
\end{tabular}
\end{table}

\section{Knowledge Base Construction Details}
\subsection{Constructing Knowledge Bases from Question Explanation}
\label{sec:kb-detail}
\begin{figure*}[h]
	\centering
	\includegraphics[width=0.95\textwidth]{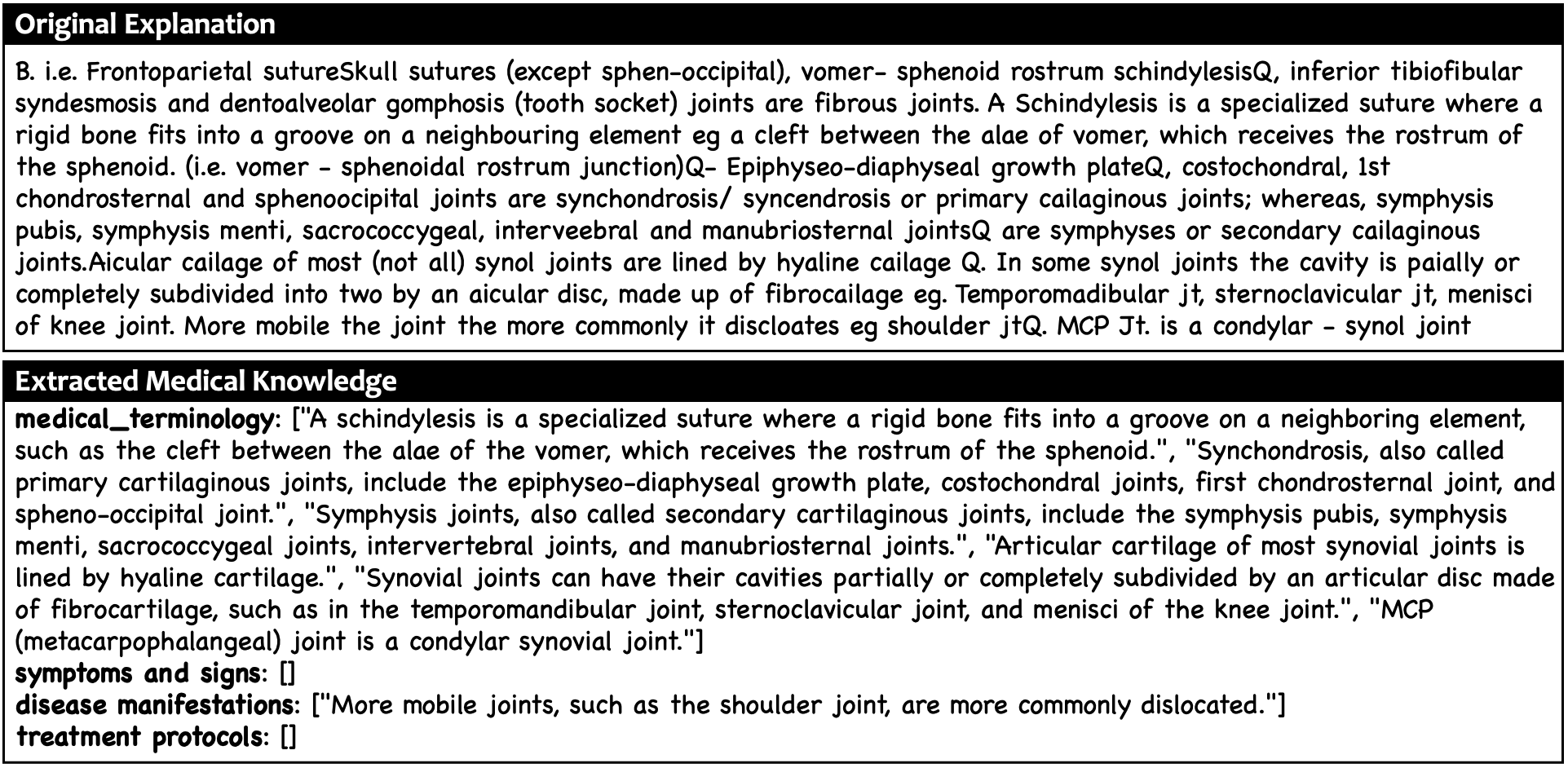}
	\caption{Example of Extracting Medical Knowledge from Explanation.}
	\label{fig:kb-sample}
\end{figure*}

To support the retrieval component of our \modelname{} framework, we construct a medical knowledge base by extracting explanatory content from available training question-answer pairs with GPT-4o. We utilize explanations from text-only datasets such as MedMCQA~\citep{pal2022medmcqa}, which contain detailed explanations from real-world medical entrance exam questions. Additionally, we also extract relevant medical knowledge from the intermediate analysis generated during our multimodal training data construction process. The employed prompt can be found in Table~\ref{tab:prompt-extract-1}.
This knowledge base encompasses medical terminology illustration, analysis of symptoms and signs, disease manifestations, and treatment protocols, as exemplified in Figure~\ref{fig:kb-sample}. By utilizing knowledge derived from highly related explanations, we ensure that the model is more likely to successfully retrieve information when querying the knowledge base during reasoning, allowing us to concentrate on evoking the model's reasoning-with-retrieval ability. Through this approach, we obtain a knowledge base, \textbf{ExpKB}, which supports the retrieve-while-reasoning training. Similarly, we construct \textbf{EC-ExpKB} from the explanations in echocardiography practice collections as a specialized knowledge base supporting model generalization to the echocardiography domain.

\begin{table}[h]
\caption{Prompt for Medical Knowledge Extraction.}
\label{tab:prompt-extract-1}
\centering
\begin{tabular}{p{0.95\linewidth}}
\toprule
\textbf{Prompt for Extracting Medical Knowledge from Explanations of Exam Questions.} \\
\midrule
\small{Extract essential medical knowledge from the following medical exam question explanation and format the output as JSON.} \\
\small{FILTER OUT these irrelevant elements: Answer designations, Citations and references, Formatting artifacts and special characters, Question numbers or identifiers} \\
\small{The JSON output should INCLUDE ONLY:} \\
\small{1. ``medical\_terminology": A list of sentences defining specialized medical terms (e.g., ``Myocardial infarction is the death of heart muscle tissue due to inadequate blood supply from coronary artery occlusion.")}\\
\small{2. ``symptoms\_and\_signs": A list of sentences describing what patients experience or what clinicians observe when having a specific disease or condition (e.g., ``Patients with myocardial infarction typically present with severe crushing chest pain.")}\\
\small{3. ``disease\_manifestations": A list of sentences describing how the disease presents, progresses, or affects the body (e.g., ``Myocardial infarction can lead to progressive heart failure over time.", ``Patients with myocardial infarction may develop life-threatening arrhythmias within 24-48 hours of the initial event.")}\\
\small{4. ``treatment\_protocols": A list of sentences describing specific medical interventions and management approaches (e.g., ``Emergency treatment for myocardial infarction includes immediate administration of aspirin 325mg chewed.", ``Primary percutaneous coronary intervention should be performed within 90 minutes of presentation if STEMI is diagnosed.")} \\
\small{Please extract only the essential medical knowledge from the following explanation using the specified JSON format. If no relevant information exists for a particular category, include that key with a empty list.
}\\
\bottomrule
\end{tabular}
\end{table}

\subsection{Applying Medical Resources as Knowledge Bases}
To assess the generalizability and practical applicability of our approach, we conducted experiments using two additional external knowledge bases derived from medical resources, 
PubMed\footnote{\href{https://ftp.ncbi.nlm.nih.gov/pub/lu/MedCPT/pubmed_embeddings}{https://ftp.ncbi.nlm.nih.gov/pub/lu/MedCPT/pubmed\_embeddings}} and StatPearls\footnote{\href{https://www.ncbi.nlm.nih.gov/books/NBK430685}{https://www.ncbi.nlm.nih.gov/books/NBK430685}}. 
While utilizing knowledge bases constructed from question explanations ensures a controlled environment for evoking reasoning-with-retrieval ability, it is also crucial to evaluate whether our proposed \modelname{} framework can effectively leverage real-world medical knowledge sources. 
For PubMed, we leverage the articles processed by \cite{jin2023medcpt}, which includes abstracts from leading medical journals across diverse specialties. 
For StatPearls, we systematically crawl the detailed explanations of medical concepts, which covers knowledge of diseases, medical conditions, interventions, diagnostic procedures, treatments, clinical guidelines, and etc. 

\subsection{Multimodal Corpus}
\label{sec:multi-kb}
We utilize a multimodal corpus with image-caption pairs for Confidence-Driven Image Retrieval. For experiments on public benchmarks, we apply the multimodal database from PubMed processed by \cite{chen2024huatuogpt-2}, PubMedVision, which includes figures with their annotated caption sourced from PubMed papers, denoted as PubMed-MMKB. For experiments on the Echocardiology domain, we apply a multimodal echocardiography expertise database, ECED~\citep{qin2025multiagent}, which includes diverse echocardiology images with their detailed descriptions.
Table~\ref{tab:kb-stat} summarizes the size of all knowledge bases used in our experiments.

\begin{table}[h]
    \caption{Size of Utilized Knowledge Bases.}
    \label{tab:kb}
    \centering
\label{tab:kb-stat}
\adjustbox{width=0.44\textwidth,center}{%
\begin{tabular}{lcc}
\toprule        
\textbf{Knowledge Base} & \textbf{Modality} & \textbf{Entries} \\
\midrule
PubMedKB & Text & 993,472 \\
StatPearlsKB & Text & 1,023,144 \\
ExpKB & Text & 58,020 \\
EC-ExpKB & Text & 1,577 \\
PubMedVision & Text \& Image & 647,031 \\
ECED & Text \& Image & 11,732 \\
\bottomrule
\end{tabular}
}
\end{table}

\section{Training Implementation Details}
\label{sec:impl}
\subsection{Prompt Design}
To guide the model to follow the predefined structure, we first craft a prompt template shown in Table~\ref{prompt-train}. The model is allowed to conduct multiple rounds of thinking and retrieval before finally reaching a decision.
\begin{table}[h]
\caption{Prompt for Evoking Reasoning-with-Retrieval}
\label{prompt-train}
\centering
\begin{tabular}{p{0.9\linewidth}}
\toprule
\textbf{Prompt Template for \modelname{}} \\
\midrule
\small{You are an experienced expert in medicine. You are given a question, an image and a list of choices. You are required to select the correct answer from the choices.
First observe the image, think about the question and each choice within} \tokensmall{<think>} \tokensmall{</think>} \small{tags. During thinking, if needed, retrieve medical knowledge using} \tokensmall{<query>} \tokensmall{</query>} \small{tags. Only one query is allowed. An external agent will retrieve information and return it within} \tokensmall{<retrieve>} \tokensmall{</retrieve>} \small{tags. 
You can use the retrieved information to continue thinking and further query if more information is needed. When you can reach a conclusion, output your answer within} \tokensmall{<answer>} \tokensmall{</answer>} \small{tags.} \\
\small{The output should be in the following format:} \\
\small{1. If you need more information, output} \tokensmall{<think>}...\tokensmall{</think>} \tokensmall{<query>}...\tokensmall{</query>} \tokensmall{<retrieve>}...\tokensmall{</retrieve>}\small{. Multiple think-query-retrieve cycles may occur.} \\
\small{2. If you can directly reach a conclusion without query, output \tokensmall{<think>}...\tokensmall{</think>} \tokensmall{<answer>}...\tokensmall{</answer>}.} \\
\bottomrule
\end{tabular}
\end{table}

\subsection{Retrieval Implementation}
\label{sec:retrieve-impl}
Our paper incorporates two types of retrieval operations: (1) text-based retrieval during the reasoning-with-retrieval process (\S~\ref{sec:text-retreive}); (2) image-based retrieval applied in Confidence-Driven Image Retrieval (\S~\ref{sec:image-retrieve}) for test-time computational scaling.
\subsubsection{Reasoning with Retrieval}
\label{sec:text-retreive}
As described in \S~\ref{sec:problem}, during the rollout process, the model first observes the image and analyzes the textual information through a reasoning process. When the model determines that its internal knowledge may be insufficient for an accurate response, it generates a query to retrieve relevant information from an external medical knowledge database. During the retrieval process, we input the query into retriever to search for related documents from the knowledge base.
For retriever, we employ the pretrained Med-CPT~\citep{jin2023medcpt} for PubMedKB and BGE-M3~\citep{chen2024m3} for the remaining KBs. Med-CPT is an information retrieval model specifically designed for the biomedicine domain, pretrained with query-article pairs from PubMed, making it ideal to retrieve from PubMed sources. BGE-M3 is a highly versatile embedding model that supports both embedding and retrieval, capable to process inputs of different scales from short sentences to long documents. We divide each item into chunks of at least 100 words for efficient indexing and retrieval. The 3 most relevant documents are retrieved for each query during the retrieval process.

\subsubsection{Confidence-Driven Image Retrieval}
\label{sec:image-retrieve}
During inference, we compensate the model with additional information from multi-mocal cases when the model's confidence in its answer is low, indicating insufficient knowledge obtained from text-based retrieval, as illustrated in~\S~\ref{sec:infer}. To retrieve similar multimodal cases, we obtain the features of input image using BioMedCLIP~\citep{zhang2023biomedclip}, a powerful foundation model pre-trained on millions of biomedical image-text pairs. To efficiently find similar cases, we compute the cosine similarity between the features of the input image and a subset of 10,000 randomly selected images from the corpus. The most similar case with the highest similarity scores alongside its corresponding caption are retrieved. We then integrate the retrieved image feature and text tokens with the original input and history contexts from the previous output, which is subsequently used for regenerating the response.
This process leverages the additional multimodal information to improve the correctness and confidence of the final output, illustrated in Algorithm~\ref{ag:img-retrieve}. 

\begin{algorithm}[h]
    \caption{Confidence-Driven Image Retrieval}
    \label{ag:img-retrieve}
    \begin{algorithmic}
        \State \textbf{Input: }Input Image and Question Pair $x=[i;t]$ , Multimodal corpus $\mathcal{D}$ containing image-caption pairs
        \State \textbf{Output: }Final Decision $y$
        \State Generate intermediate turns of think, query, retrieve  $h_n,k_n \leftarrow\pi_\theta(x,h_{1:n-1},k_{1:n-1})$
        \State Generate output $\hat{y}\leftarrow\pi_\theta(\cdot\mid x,h_{1:n},k_{1:n})$ 
        \State Calculate answer confidence $\eta\leftarrow\pi_{\theta}(\hat{y}_{\tau+1}^{\text{answer}}\mid x,\hat{y}_{\leq\tau},h_{1:n},k_{1:n})$
        \If{$\eta<0.8$} \textcolor{darkblue}{\Comment{low confidence}}
        \State Conduct image retrieval $(i_{\text{sim}},t_{\text{sim}})\leftarrow\mathcal{k}_{\text{sim}}(i,\mathcal{D})$ 
        \State$y\leftarrow\pi_\theta(\cdot\mid x,h_{1:n},k_{1:n},i_{\text{sim}},t_{\text{sim}})$ \textcolor{darkblue}{\Comment{incorporate the retrieved multimodal case}}
        \Else \textcolor{darkblue}{\Comment{high confidence}}
        \State Apply the confident answer $y\leftarrow\hat{y}$
        \EndIf
    \end{algorithmic}
\end{algorithm}

\subsection{Implementation Settings}
All experiments are conducted on 8 H800 GPUs using SWIFT~\citep{zhao2025swift} as the training framework. For GRPO training, we perform full-parameter fine-tuning with a learning rate of $1e^{-6}$ and $\beta$ set as $1e^{-3}$, optimized using DeepSpeed with Zero-2 Configuration. 

To ensure a fair comparison, we modify the baseline setup for the Agentic Search MLLMs. Instead of applying their original online search environments, the baselines were configured to use the same knowledge base and retriever as our proposed framework.




\subsection{Training Dynamics}
\begin{figure*}[h]
	\centering
	\includegraphics[width=1.0\textwidth]{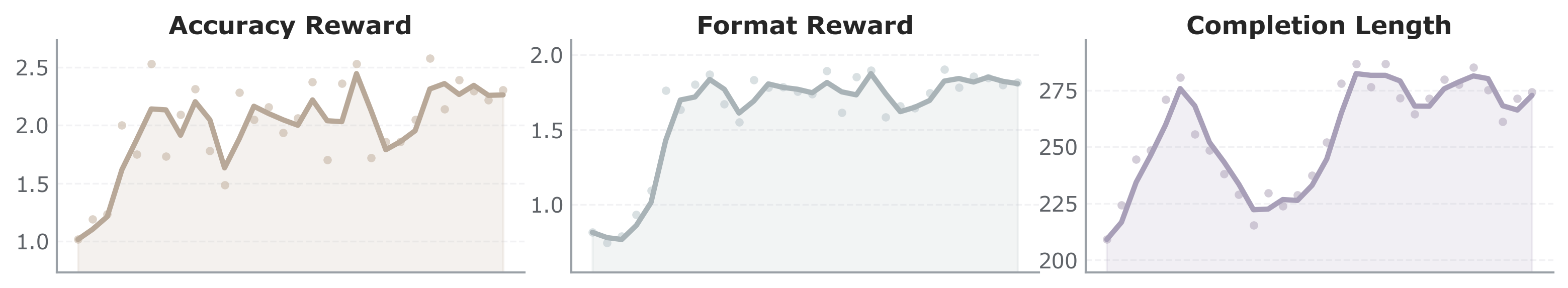}
	\caption{Rewards and Response Length across Training Iterations of the First Stage. The rewards are plotted every five steps.}
	\label{fig:reward-text}
\end{figure*}

In the first stage of training, we concentrate on eliciting the model’s reasoning-with-retrieval ability with text-only question-answer pairs through accuracy and format rewards. The evolution of rewards and completion length over training steps are shown in Figure~\ref{fig:reward-text}.

\begin{figure*}[h]
	\centering
	\includegraphics[width=1.0\textwidth]{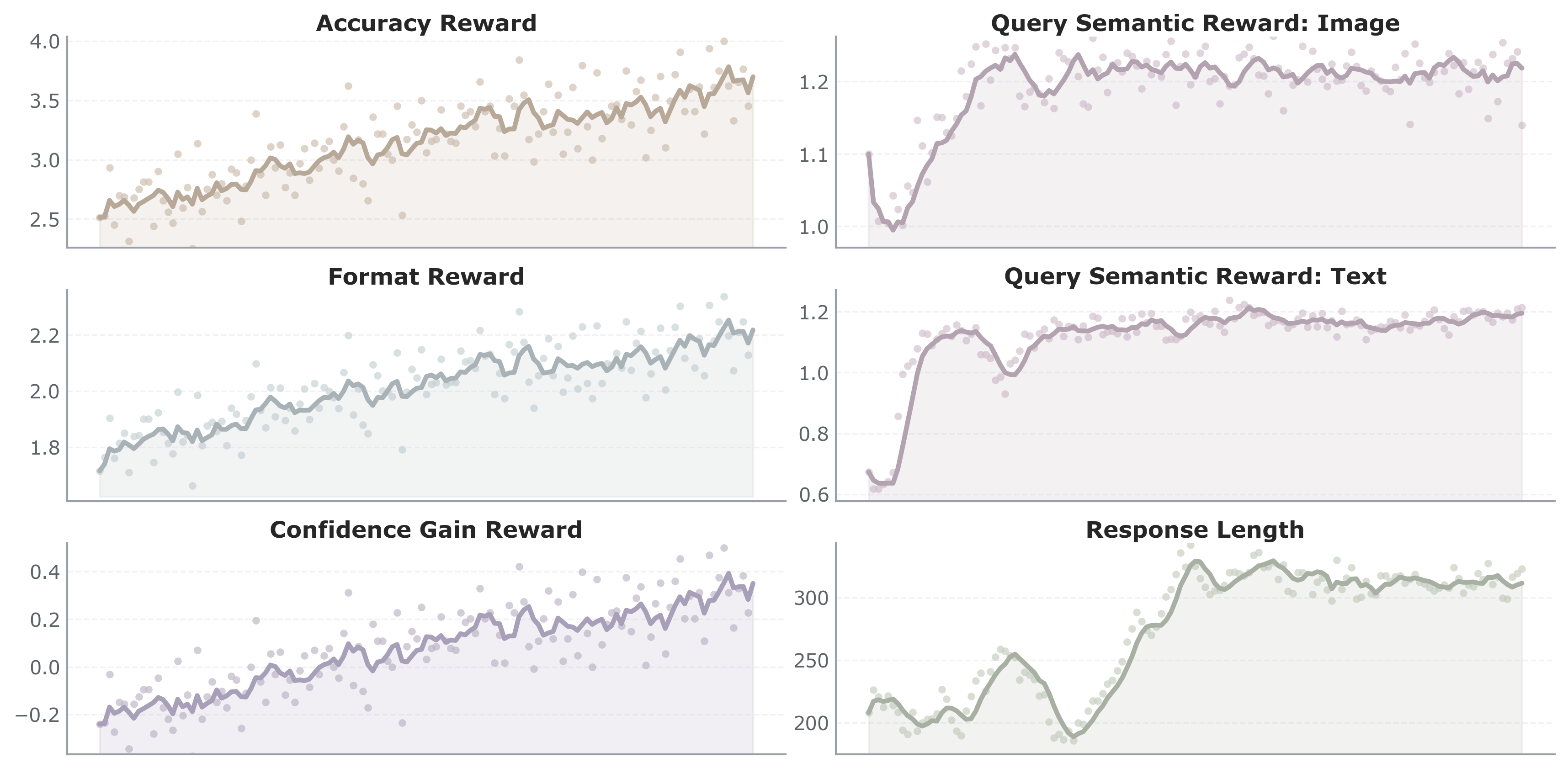}
	\caption{Rewards and Response Length across Training Iterations of the Second Stage. The rewards are plotted every five steps.}
	\label{fig:reward}
\end{figure*}

In the second stage of training, all four reward types proposed are incorporated to enhance the effectiveness of query formulation and relevance of retrieved knowledge. Figure~\ref{fig:reward} demonstrates the reward and response length change across training iterations.

\section{Supplementary Experiments}
\label{sec:supl-exp}
\subsection{Comparison with Reasoning Data Supervised Fine-tuning}
To validate the effectiveness of our two-stage training strategy, we compared with employing text-only supervised fine-tuning (SFT) instead of GRPO training in the first stage, while maintaining multimodal GRPO training in the second stage. For the SFT data, since our text-only GRPO training data does not contain explicit reasoning chains necessary for supervised fine-tuning, we employed two publicly released medical reasoning datasets derived from the same medical QA sources,he MedQA-USMLE~\citep{jin2021disease} and MedMCQA~\citep{pal2022medmcqa}, as we used for GRPO training. As shown in Table~\ref{tab:sft}, we attribute this performance drop to the model's tendency to overfit reasoning templates during SFT training without adequately understanding the inherent medical knowledge, which appears to constrain the model's ability to dynamically retrieve and synthesize relevant information. Future directions could explore developing more adaptive approaches to cold-start data construction in the reasoning-with-retrieval scenario.
\begin{table}[h]
    \caption{Comparison with Applying Supervised Fine-tuning in the First Stage to Warm up Reasoning.}
    \label{tab:sft}
    \centering
    \begin{adjustbox}{width=0.95\textwidth}
        \begin{tabular}{llcccc}
        \toprule        
        \textbf{First Stage} & \textbf{Second Stage} &\textbf{MedXpertQA-MM} & \textbf{MMMU-H\&M} & \textbf{MMMU-Pro-H\&M} & \textbf{ECBench} \\
        \midrule
        Med-o1 SFT & Multimodal GRPO& 24.1 & 55.3 & 33.6 & 39.7 \\
        MedReason SFT & Multimodal GRPO & 26.1 & 52.0 & 36.7 & 39.8 \\
        Text-only GRPO & Multimodal GRPO & 27.2 & 65.5 & 43.7 & 51.1 \\
        \bottomrule
        \end{tabular}
    \end{adjustbox}
\end{table}

\subsection{Integrating Different Knowledge Bases}
To assess the generalizability and practical applicability of our approach toward different Knowledge Bases (KBs), we conduct experiments with different KBs described in~\ref{sec:kb-detail} and present the results in Table~\ref{tab:diffkb}. 
The results demonstrate that the model achieves the best performance with our constructed KB derived from exam explanations on general medical QA tasks, while showing optimal results in echocardiography domain with the specialized echocardiography KB. This is probably because our constructed KBs contain fewer but more targeted entries as shown in the KB statistics (Table~\ref{tab:kb-stat}). This enables the retriever to more effectively identify relevant contents for each query.
In contrast, larger-scale KBs, while containing more comprehensive knowledge, present greater challenges for the retrieval systems. This suggests that for knowledge bases with substantial scale and diversity, further developing a more sophisticated retrieval mechanism becomes essential. 

\begin{table}[h]
    \caption{Results of Applying External Knowledge from Different KBs.}
    \label{tab:diffkb}
    \centering
    \begin{adjustbox}{width=0.75\textwidth}
        \begin{tabular}{lcccc}
        \toprule
        \textbf{} & \textbf{MedXpertQA-MM} & \textbf{MMMU-H\&M} & \textbf{MMMU-Pro-H\&M} & \textbf{ECBench} \\
        \midrule
        PubMedKB & 26.4 & 61.3 & 41.3 & 49.0 \\
        StatpearlsKB & 26.6 & 61.4 & 42.0 & 49.3 \\
        ExpKB & \textbf{27.2} & \textbf{65.5} & \textbf{43.7} & 48.3 \\
        EC-ExpKB & 25.4 & 62.1 & 40.2 & \textbf{51.1} \\
        \bottomrule
        \end{tabular}
    \end{adjustbox}
\end{table}

\subsection{Experiments on Traditional Medical VQA Datasets}
We also report the accuracy on traditional medical VQA datasets, SLAKE~\citep{liu2021slake}, VQA-RAD~\citep{lau2018dataset}, PathVQA~\citep{he2020pathological}, PMC-VQA~\citep{zhang2023pmc}, and OmniMedVQA~\citep{hu2024omnimedvqa}, in Table~\ref{tab:result-vqa}. Note that \textit{our model is not trained with any data from the training split of these target datasets}. Compared with the base model QwenVL-2.5, the performance gains achieved by our approach are consistent but not substantial. This can possibly be attributed to the difference in dataset complexity and the requirements of reasoning abilities. These datasets contain large amounts of questions that can be directly inferred from images without requiring complicated reasoning or external knowledge, and thus do not necessitate the reasoning-with-retrieval capability our model enhances.
\begin{table}[h]
    \caption{Performance on Traditional Medical VQA Datasets.}
    \label{tab:result-vqa}
    \centering
    \begin{adjustbox}{width=0.75\textwidth}
        \begin{tabular}{lccccc}
        \toprule        
        \textbf{Model} & \textbf{SLAKE} & \textbf{VQA-RAD} & \textbf{PathVQA} & \textbf{PMC-VQA} & \textbf{OmniMedVQA} \\
        \midrule
        QwenVL-2.5 & 74.0 & 73.1 & 64.8 & 53.0 & 63.9 \\
        \modelname{} & 75.5 & 75.4 & 69.9 & 53.5 & 73.1 \\
        \bottomrule
        \end{tabular}
    \end{adjustbox}
\end{table}

\subsection{Evaluating Reasoning Path with the LLM-as-a-Judge Strategy}
We further apply an LLM-as-a-Judge strategy to evaluate the quality of the reasoning path. Specifically, we apply Qwen3-32B~\citep{yang2025qwen3} to assess the generated reasoning path on ECBench from four aspects: (1) Coherence: assesses if the generated reasoning is coherent, grammatically correct, and fluent; (2) Factuality: assess if the reasoning is grounded on correct and verifiable clinical knowledge; (3) Relevance: assess if the reasoning directly addresses the problem posed in the question and logically supports the given answer; (4) Conciseness: assess if the reasoning avoid redundant steps and irrelevant details. A score between 0 and 5 is assigned to each aspect.
Results are shown in Table~\ref{tab:llm-judge}.
\begin{table}[h]
    \caption{Reasoning Path Evaluation with LLM-as-a-Judge Strategy.}
    \label{tab:llm-judge}
    \centering
    \begin{adjustbox}{width=0.85\textwidth}
        \begin{tabular}{lccccc}
        \toprule        
        \textbf{Model} & \textbf{Coherence} & \textbf{Factuality} & \textbf{Relevance} & \textbf{Conciseness} & \textbf{Average} \\
        \midrule
        MedVLM-R1~\citep{pan2025medvlm} & 3.2 & 1.7 & 2.4 & 3.4 & 2.7 \\
        Med-R1~\citep{lai2025med} & 3.7 & 2.1 & 2.4 & 4.0 & 3.1 \\
        Chiron-o1~\citep{sun2025enhancing} & 4.2 & 2.8 & 3.5 & 4.0 & 3.6 \\
        \modelname{} (Ours) & \textbf{4.4} & \textbf{3.1} & \textbf{3.8} & \textbf{4.4} & \textbf{3.9} \\
        \bottomrule
        \end{tabular}
    \end{adjustbox}
\end{table}


\section{Case Studies}
\label{sec:case}
\subsection{Case Studies for Reasoning-with-Retrieval Process}
\begin{figure*}[h]
	\centering
	\includegraphics[width=0.9\textwidth]{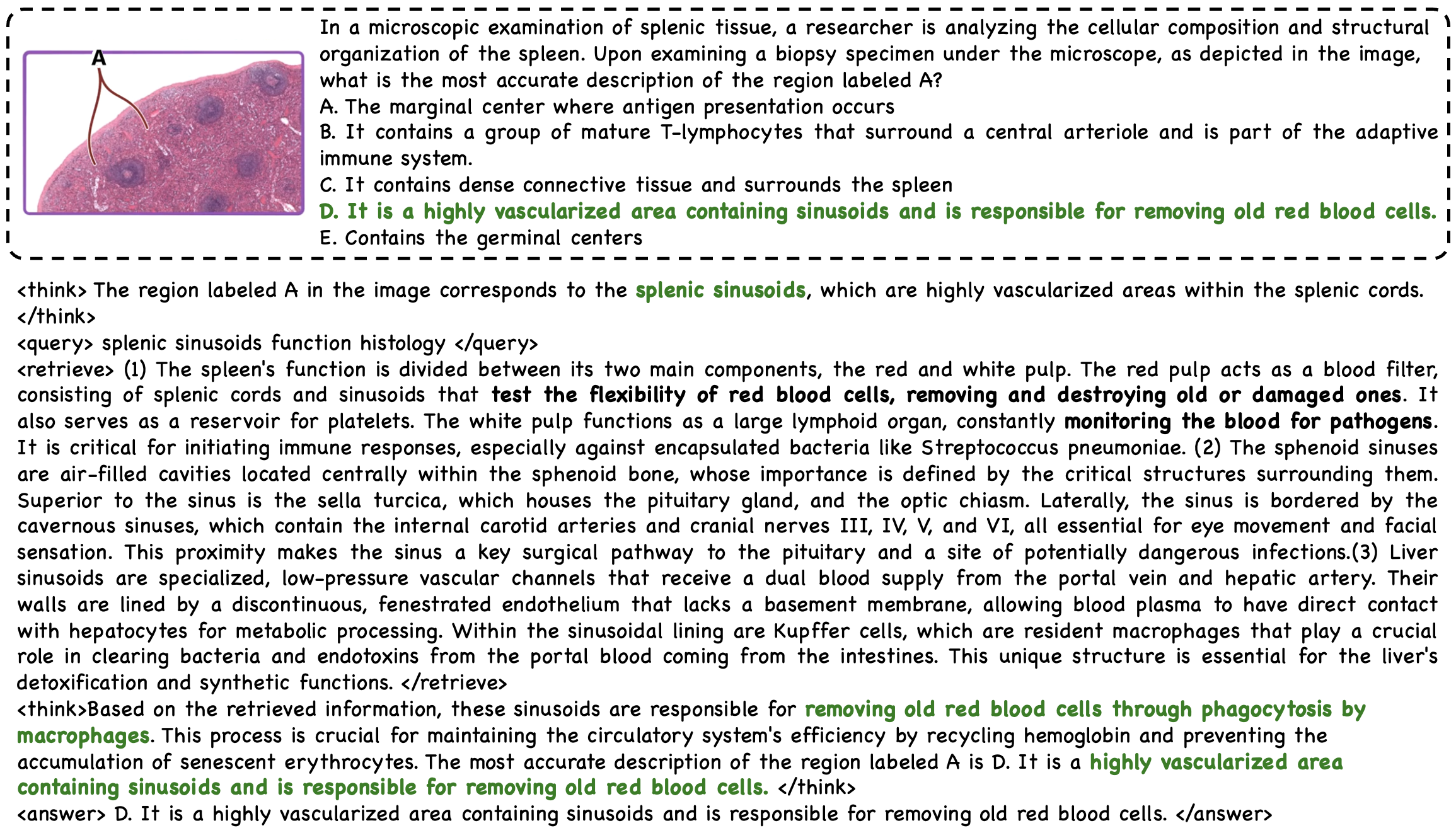}
	\caption{Example of \modelname{}'s Rollout Process: Observation-Driven Retrieval.}
	\label{fig:case-1}
\end{figure*}

\begin{figure*}[h]
	\centering
	\includegraphics[width=0.9\textwidth]{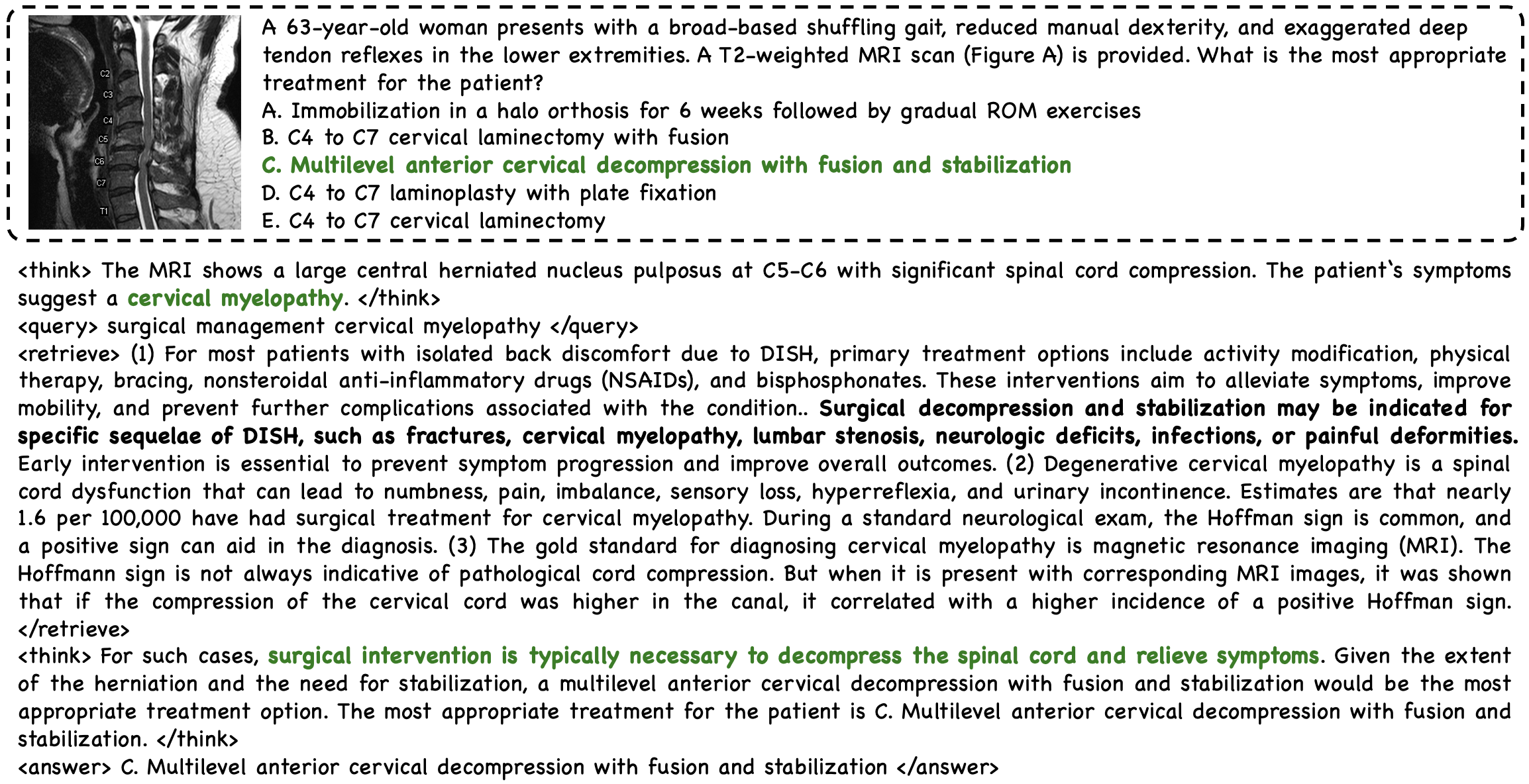}
	\caption{Example of \modelname{}'s Rollout Process: Hypothesis-Driven Retrieval.}
	\label{fig:case-2}
\end{figure*}

\begin{figure*}[h]
	\centering
	\includegraphics[width=0.9\textwidth]{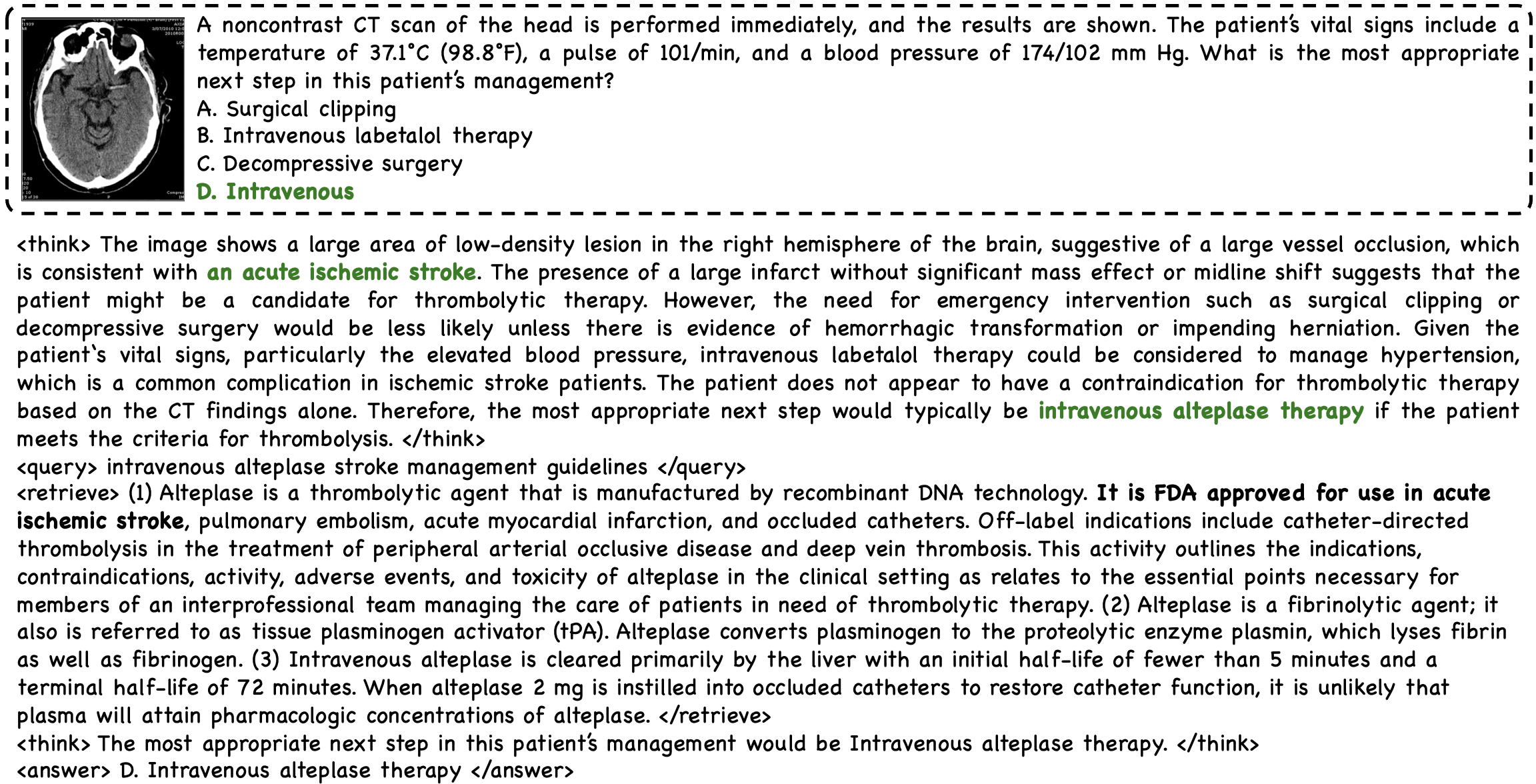}
	\caption{Example of \modelname{}'s Rollout Process: Confirmation-Driven Retrieval.}
	\label{fig:case-3}
\end{figure*}

We demonstrate several case studies of the rollout process generated with our proposed reasoning-with-retrieval framework \modelname{}, revealing how the retrieval mechanism could evolve with clinical reasoning. Figure~\ref{fig:case-1} provides an example of \textit{observation-driven retrieval} when model encounter unfamiliar visual observations in the medical scan, such as the anatomical structure ``splenic sinusoids". This triggers the model to retrieve relevant knowledge, including definitions and physiological functions, which may potentially help interpret the image and reach the final conclusion. \textit{Hypothesis-driven retrieval} emerges as shown in Figure~\ref{fig:case-2} when the model can relate visual findings with a specific clinical symptom to formulate a preliminary diagnosis, but requires additional information on treatment pathways, prognostic assessments, and so forth. When the model possesses sufficient internal knowledge to reach clinical conclusions as an example in Figure~\ref{fig:case-3}, \textit{confirmation-driven retrieval} is also helpful to ensure adherence to clinical protocols and ground reasoning in authoritative guidelines.

\subsection{Case Studies for Confidence-Driven Image Re-retrieval}
\begin{figure}[h]
	\centering
	\includegraphics[width=0.9\textwidth]{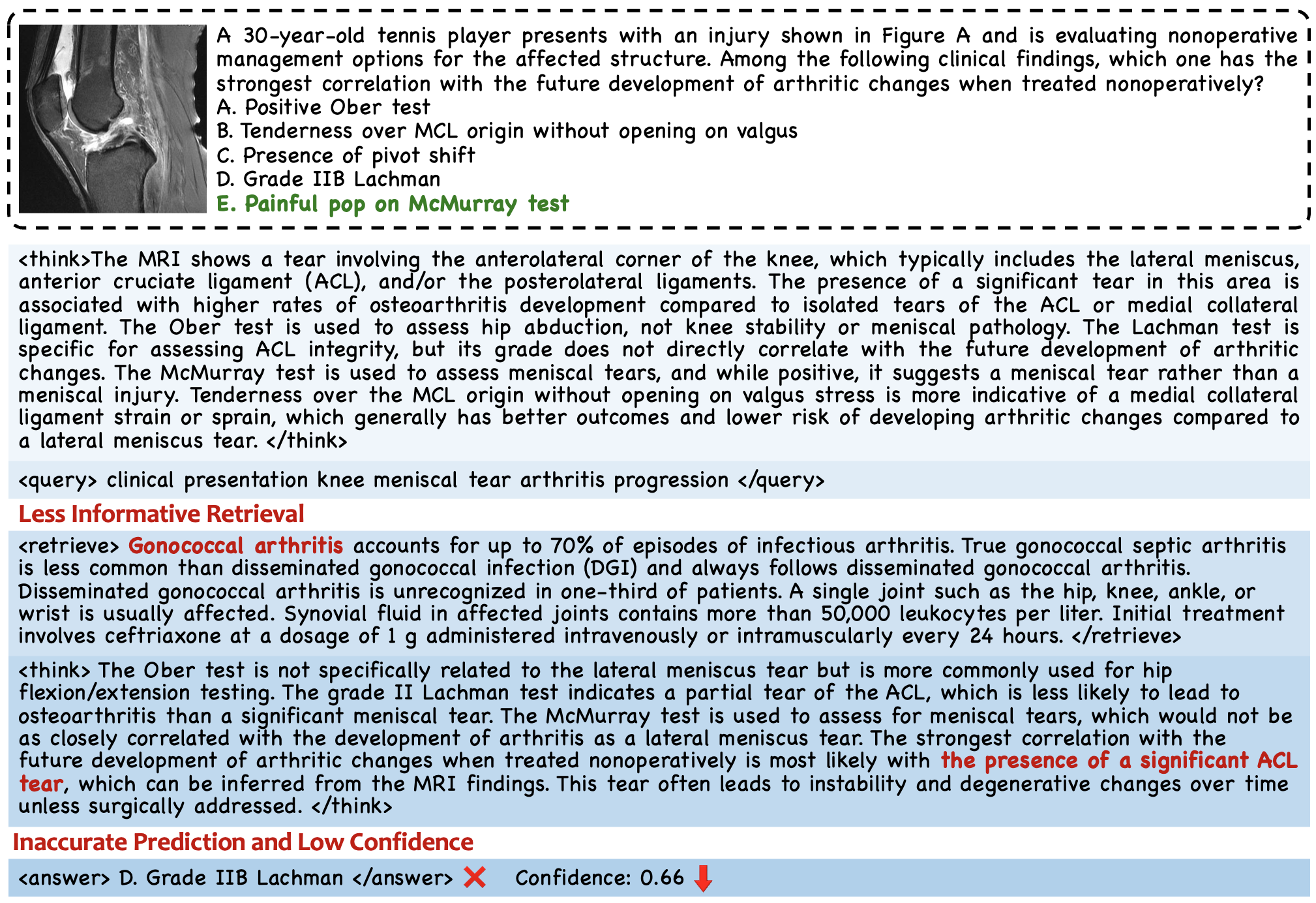}
	\caption{Example of Model Rollout Process without \textsc{CDIR}.}
	\label{fig:before-cdir}
\end{figure}

\begin{figure}[h]
	\centering
	\includegraphics[width=0.9\textwidth]{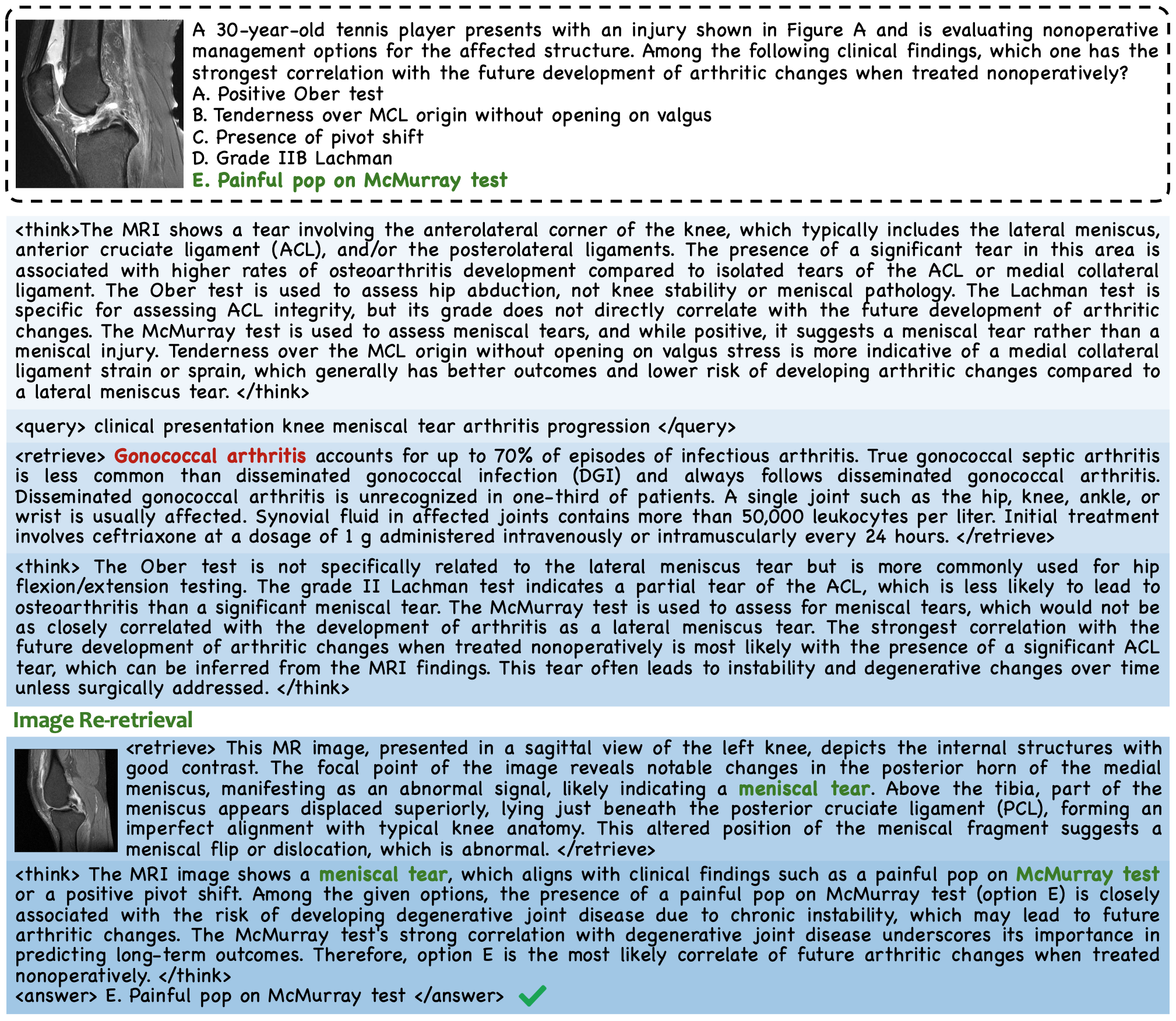}
	\caption{Example of Model Rollout Process with \textsc{CDIR}.}
	\label{fig:after-cdir}
\end{figure}

To illustrate the efficacy of the Confidence-Driven Image Re-retrieval (\textsc{CDIR}) mechanism, we present a case study in Figure~\ref{fig:before-cdir} and Figure~\ref{fig:after-cdir}, which demonstrate the rollout process with and without \textsc{CDIR}. Initially, while the model formulated an appropriate and clinically relevant query ``knee meniscal tear arthritis progression" based on the input image and clinical context, the knowledge retrieval process failed to return helpful information from the knowledge base. The retrieval ineffectiveness resulted in incorrect decision-making, as it failed to help the model differentiate the clinical significance of an ACL injury from a meniscal tear in this context. This is potentially reflected by the model's low confidence, which fell below the predefined threshold $\lambda=0.8$, so \textsc{CDIR} mechanism is triggered. The model then performed a re-retrieval by searching the multimodal database for images with high similarity. In this example, a similar case depicting a meniscal tear is retrieved, accompanied by a descriptive caption. By integrating this highly pertinent multimodal information, the model was able to rectify its initial reasoning and correctly identify the correct answer. This case demonstrates that \textsc{CDIR} could overcome situations where the initial knowledge base is insufficient, thereby enhancing diagnostic accuracy.

\clearpage

\subsection{Ablative Case Studies}
\begin{figure*}[h]
	\centering
	\includegraphics[width=1.0\textwidth]{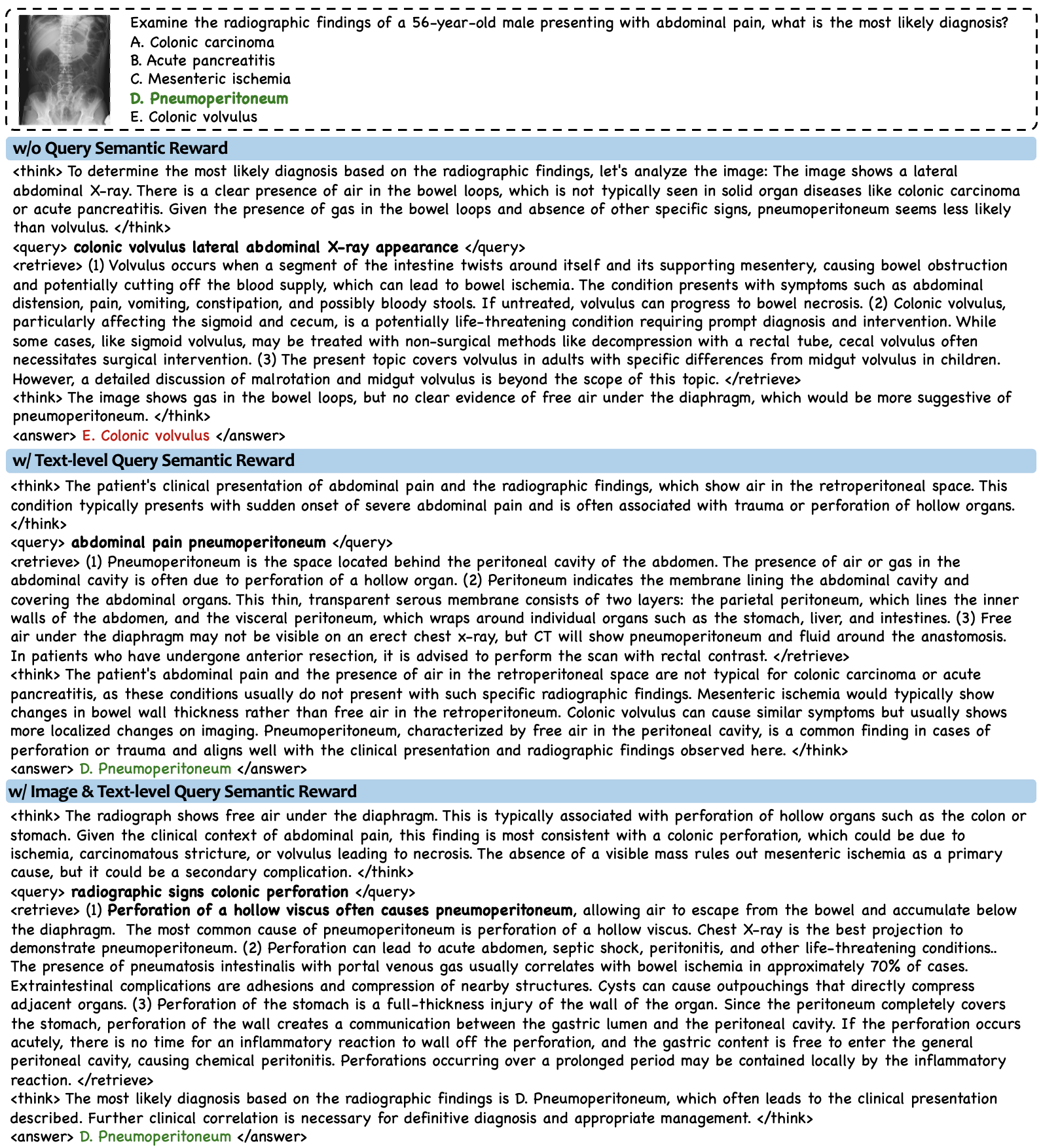}
	\caption{Ablative Case Studies of the Query Semantic Reward.}
	\label{fig:abl-case}
\end{figure*}

We provide ablative case studies in Figure~\ref{fig:abl-case} to demonstrate the effectiveness of the query semantic reward. While applying only the query semantic reward for textual alignment successfully helps the model identify the correct answer, the generated reasoning reveals an uninterpretable logical process. The generated query, ``abdominal pain pneumoperitoneum" is a direct combination of the patient's symptom from the question and one of the potential answers, failing to explain why the specific diagnosis was investigated. 
In contrast, the model guided by the query semantic reward for multimodal alignment demonstrates a more effective and interpretable reasoning process. The multimodal alignment incentivizes queries that not only originate from a key visual observation but also relate it to textually relevant knowledge. 

\subsection{Failure Case Analysis}
\begin{figure*}[h]
	\centering
	\includegraphics[width=1.0\textwidth]{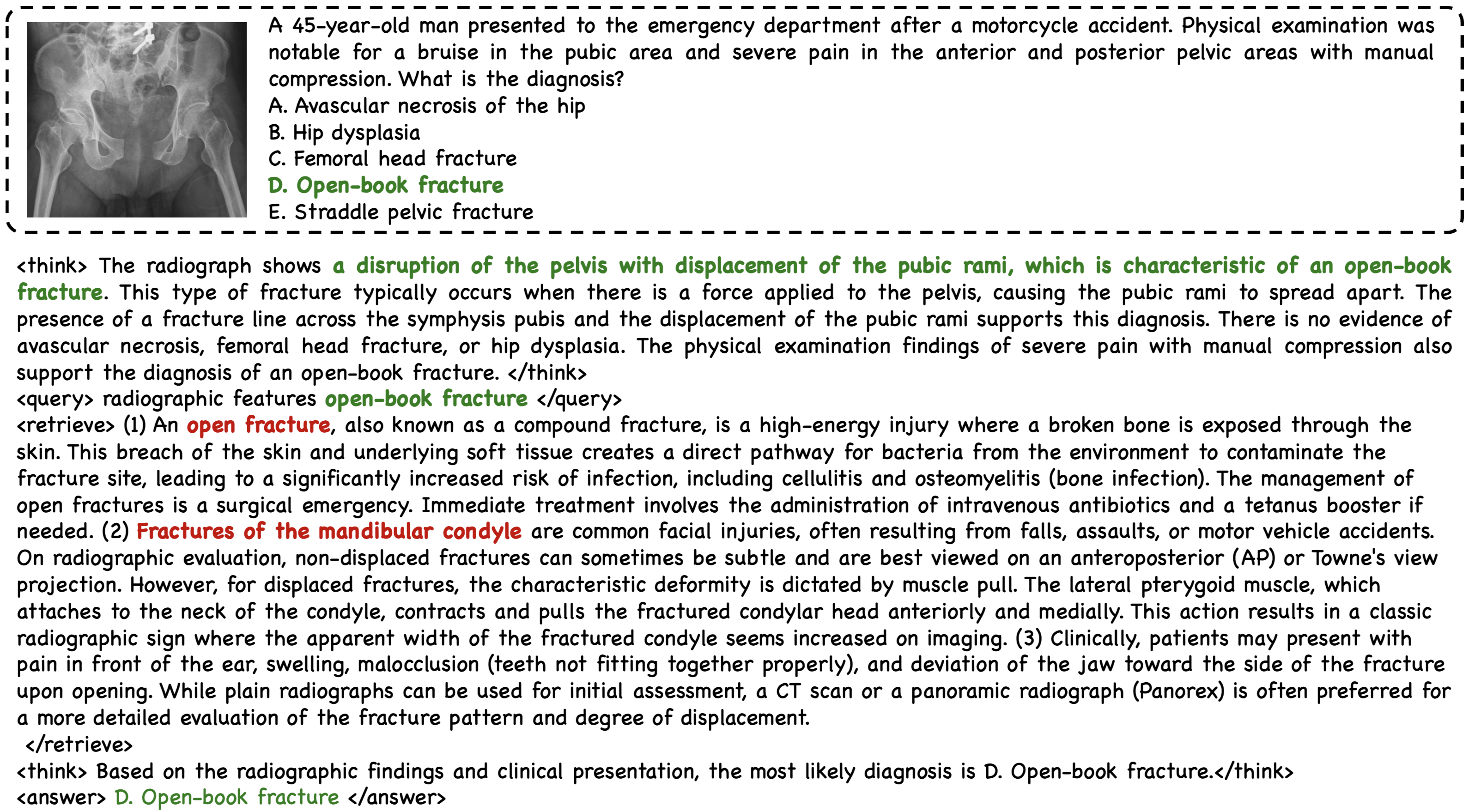}
	\caption{Example of Failure Case: Retrieval Failure}
	\label{fig:fail-case-1}
\end{figure*}

\begin{figure*}[h]
	\centering
	\includegraphics[width=1.0\textwidth]{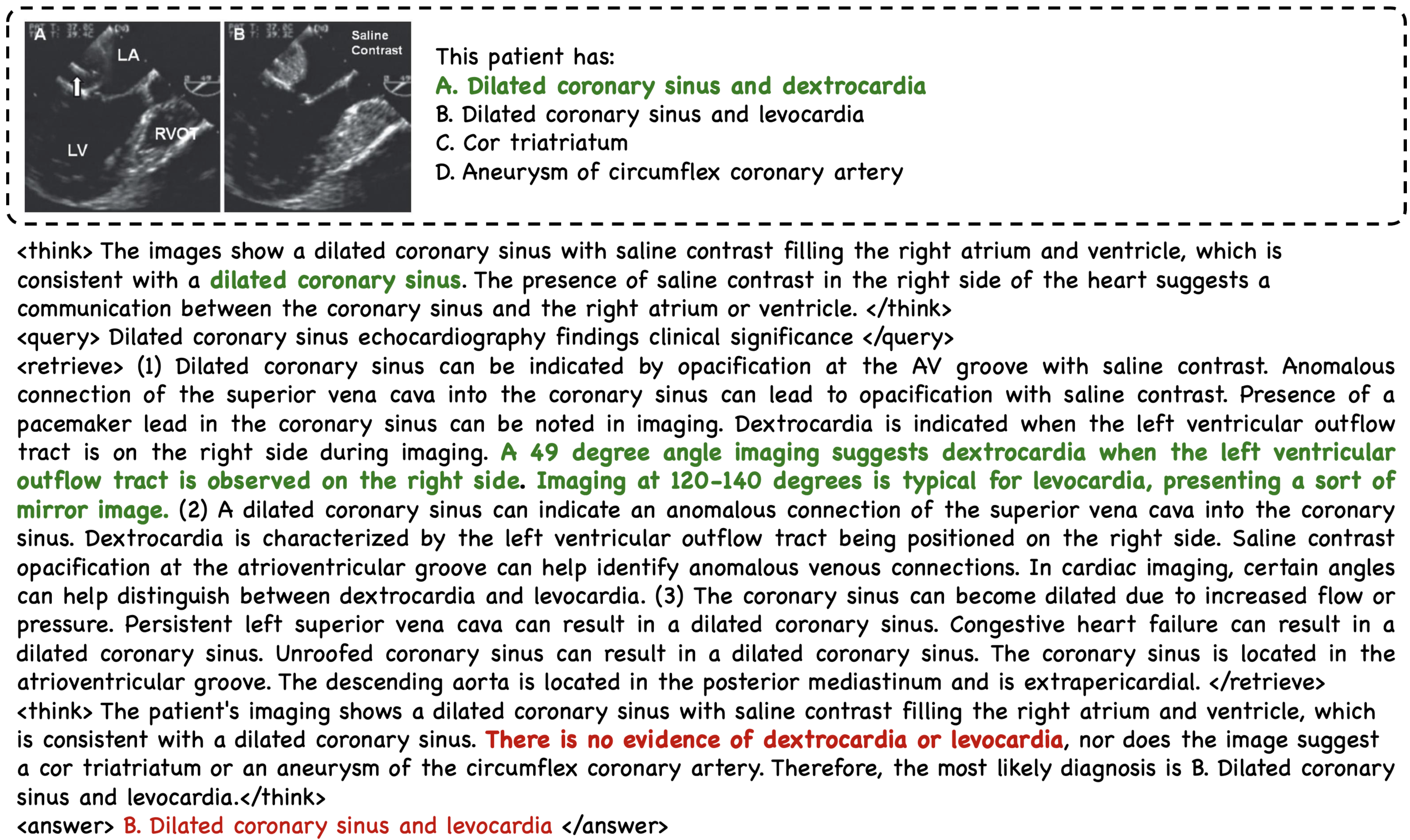}
	\caption{Example of Failure Case: Integration Failure}
	\label{fig:fail-case-2}
\end{figure*}

\begin{figure*}[h]
	\centering
	\includegraphics[width=1.0\textwidth]{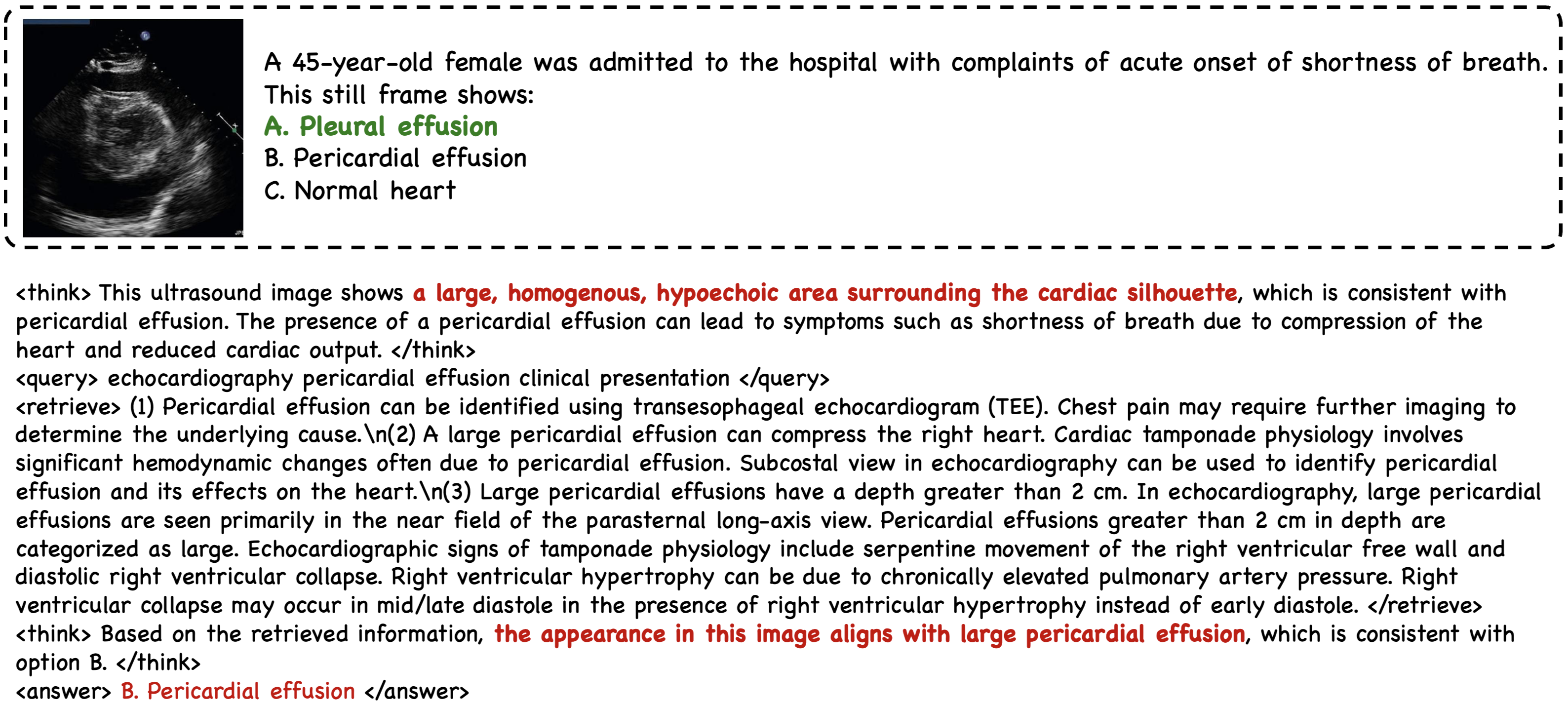}
	\caption{Example of Failure Case: Perceptual Failure}
	\label{fig:fail-case-3}
\end{figure*}

To gain a deeper understanding of the behavior and capabilities of our proposed \modelname{}, we conduct additional analysis of failure cases to identify potential limitations and illuminate future directions for enhancement. We generalize the failures into three categories: (1) \textit{Retrieval Failure} (Figure~\ref{fig:fail-case-1}): the retriever returns irrelevant information despite an effective query; (2) \textit{Integration Failure} (Figure~\ref{fig:fail-case-2}): the model fails to incorporate the helpful retrieved knowledge into the reasoning process, adhering to the original reasoning path. (3) \textit{Perceptual Failure} (Figure~\ref{fig:fail-case-3}): the model's insufficient internal knowledge about specific conditions leads to the misinterpretation of the medical scan. These cases indicate that our proposed framework can extend the model's upper bound by encouraging the model to actively retrieve external knowledge. However, enriching the base MLLM with more extensive medical knowledge while enhancing the retriever's ability to identify relevant information from extensive knowledge is also crucial to unlock the full potential of reasoning-with-retrieval capability.

\section{Supplementary Related Work}
\subsection{Medical Multimodal Large Language Models}
Current advances in MLLMs have accelerated their application in medical settings to assist diagnostic procedures. 
Recent developments fall into two main directions. Specialist models are designed for specific medical modalities such as radiology~\citep{wu2023towards} and pathology~\citep{lu2024multimodal}, or targeted at specialized tasks~\citep{bannur2024maira,ding2024hia}. Conversely, generalist models are designed for broader medical assistance across different modalities~\citep{zhang2024generalist,wang2024interpretable,yang2025multi}. 

However, these models directly provide diagnostic outputs without detailed explanations, making it difficult for clinicians to interpret and follow. This lack of transparency has driven the development of medical reasoning models, which explicitly output the thinking process for decision. Several works focused on curating Long-CoT data to teach medical models step-by-step reasoning~\citep{chen2024huatuogpt,huang2025o1,wu2025medreason}, 
and recent efforts have applied reinforcement learning to iteratively improve reasoning quality through reward-based training~\citep{xu2025medground,su2025gmai,liu2025x,mu2025elicit}. Despite these improvements, a fundamental challenge remains. Both Long CoT and RL-based approaches rely heavily on the models' internal knowledge in the inference stage, which would generate reasoning processes that contain unreliable medical information especially when encountering data out of training scope, potentially compromising clinical decision-making.

\subsection{Agentic Search and Retrieval}
Researchers have explored integrating retrieval into the reasoning process to enable mutual enhancement where the retrieved information can support reasoning, while reasoning helps generate more effective queries.
Self-RAG~\citep{asai2024self} adaptively and iteratively retrieves relevant information when required, improving both generation quality and self-reflection capabilities. \cite{jeong2024improving} adapts similar insights into the medical domain, allowing the model to emulate medical expert behavior. 
Recent works have introduced Reinforcement Learning (RL) to stimulate retrieval abilities during reasoning, reducing the dependence on large-scale annotated data for supervised fine-tuning. Search-R1~\citep{jin2025search} and R1-Searcher~\citep{song2025r1} propose reinforcement learning frameworks that integrate external search capabilities into reasoning systems, enabling models to augment their internal knowledge with dynamically retrieved information during decision-making.
DeepResearcher~\citep{zheng2025deepresearcher} incorporates direct interaction with real-world search engines. ZeroSearch~\citep{sun2025zerosearch} addresses unpredictable document quality and prohibitive API costs by leveraging a trained LLM to simulate real-time search. \cite{ding2025promed,zheng2025end,yu2025medreseacher} have extended the agentic RAG reasoning into medical LLMs, allowing proactive search for related articles, textbooks, and clinical guidelines.

In the multimodal domain, Visual-ARFT~\citep{liu2025visual2} proposed an effective framework that equips MLLMs with agentic capabilities of text-based searching. MMSearch-R1~\citep{wu2025mmsearch} allows the model to perform adaptive multimodal retrieval from real-world Internet sources. 
VRAG-RL~\citep{wang2025vrag} further enhances multimodal retrieval with on-demand trigger of visual actions like cropping and scaling the original input image.
Despite these advances, there remains a critical gap in exploring multimodal reasoning with adaptive retrieval in the medical domain, where the retrieved knowledge often supports differential diagnosis reasoning among multiple hypotheses, rather than seeking a single correct answer. This requires the model to aware of anatomical structures and pathological patterns that general agentic search models could not achieve. 

\section{Usage of Large Language Models}
Large Language Models (LLMs) were used to polish languages and refine structures in paper writing. Additionally, LLMs also assist in data curation process as described in Appendix~\ref{sec:train-data-construct} and~\ref{sec:ecbench}.
\end{document}